\documentclass[nohyperref]{article}

\usepackage{microtype}
\usepackage{graphicx}
\usepackage{subfigure}
\usepackage{booktabs} %

\PassOptionsToPackage{hyphens}{url}
\usepackage{hyperref}
\hypersetup{colorlinks=true,breaklinks=true}

\usepackage[accepted]{icml2022}

\usepackage{amsmath}
\usepackage{amssymb}
\usepackage{mathtools}
\usepackage{amsthm}

\usepackage{enumitem}
\usepackage{physics}
\usepackage[nice]{nicefrac}

\usepackage{amsmath,amsfonts,bm,amsthm,dsfont}

\def\eqref#1{equation~\ref{#1}}

\def\1{\bm{1}}
\def\0{\bm{0}}

\def\vc{{\bm{c}}}

\def\vp{{\bm{p}}}

\def\vx{{\bm{x}}}

\def\vz{{\bm{z}}}

\def\mA{{\bm{A}}}

\def\mI{{\bm{I}}}

\def\mM{{\bm{M}}}

\def\mP{{\bm{P}}}

\def\mT{{\bm{T}}}

\def\mW{{\bm{W}}}
\def\mX{{\bm{X}}}

\def\mZ{{\bm{Z}}}

\DeclareMathAlphabet{\mathsfit}{\encodingdefault}{\sfdefault}{m}{sl}
\SetMathAlphabet{\mathsfit}{bold}{\encodingdefault}{\sfdefault}{bx}{n}

\newcommand{\E}{\mathbb{E}}

\DeclareMathOperator*{\argmax}{arg\,max}

\DeclareMathOperator{\sign}{sign}

\newcommand{\PP}{\mathbb P}

\usepackage{tcolorbox}
\definecolor{grey}{rgb}{0.33, 0.33, 0.33}

\newcommand{\squishlist}{
\begin{list}{{{\small{$\bullet$}}}}
{\setlength{\itemsep}{1pt}      \setlength{\parsep}{1pt}
\setlength{\topsep}{-2pt}       \setlength{\partopsep}{0pt}
\setlength{\leftmargin}{1em} \setlength{\labelwidth}{1em}
\setlength{\labelsep}{0.5em} } }
\newcommand{\squishend}{  \end{list}  }

\makeatletter
\renewcommand*\env@matrix[1][*\c@MaxMatrixCols c]{%
  \hskip -\arraycolsep
  \let\@ifnextchar\new@ifnextchar
  \array{#1}}
\makeatother

\usepackage{listings}

\usepackage{graphicx,subfigure,epsfig}
\usepackage{comment}
\usepackage{xcolor,enumerate}

\usepackage{threeparttable}

\usepackage{multirow}

\usepackage{tcolorbox}
\definecolor{grey}{rgb}{0.33, 0.33, 0.33}

\usepackage{algorithmic}
\usepackage{algorithm}
\usepackage{enumitem}

\ifodd 0
\newcommand{\rev}[1]{{\color{blue}#1}}
\newcommand{\yl}[1]{\textbf{\color{red}(Yang: #1)}}
\newcommand{\zzw}[2]{\textbf{\color{blue}(Zhaowei: #1)}{\color{blue}~#2}}
\newcommand{\jl}[1]{\textbf{\color{magenta}(Jialu: #1)}}

\newcommand{\clar}[1]{\textbf{\color{green}(NEED CLARIFICATION: #1)}}

\else
\newcommand{\rev}[1]{#1}
\newcommand{\com}[1]{}
\newcommand{\clar}[1]{}
\newcommand{\response}[1]{}
\newcommand{\yl}[1]{}
\newcommand{\zzw}[2]{}
\newcommand{\jl}[1]{}
\fi

\newcommand{\algcom}[1]{\textsl{\color{blue}{\footnotesize #1}}}

\usepackage{lipsum}

\usepackage[capitalize,noabbrev]{cleveref}

\theoremstyle{plain}
\newtheorem{theorem}{Theorem}[section]

\newtheorem{lemma}[theorem]{Lemma}
\newtheorem{corollary}[theorem]{Corollary}
\theoremstyle{definition}
\newtheorem{definition}[theorem]{Definition}

\theoremstyle{remark}
\newtheorem{remark}[theorem]{Remark}

\usepackage[textsize=tiny]{todonotes}

\newcommand{\setalglineno}[1]{%
  \setcounter{ALC@line}{\numexpr#1-1}}

\icmltitlerunning{Label Noise Transition Matrix Estimation for Tasks with Lower-Quality Features}

\begin{document}

\twocolumn[
\icmltitle{Beyond Images: \\
\hspace{-10pt}Label Noise Transition Matrix Estimation for Tasks with Lower-Quality Features}

\icmlsetsymbol{equal}{*}

\begin{icmlauthorlist}
\icmlauthor{Zhaowei Zhu}{UCSC}
\icmlauthor{Jialu Wang}{UCSC}
\icmlauthor{Yang Liu}{UCSC}
\end{icmlauthorlist}

\icmlaffiliation{UCSC}{Department of Computer Science and Engineering, University of California, Santa Cruz, CA, USA}

\icmlcorrespondingauthor{Yang Liu}{yangliu@ucsc.edu}

\icmlkeywords{Machine Learning, ICML}

\vskip 0.3in
]

\printAffiliationsAndNotice{}  %

\begin{abstract}
The label noise transition matrix, denoting the transition probabilities from clean labels to noisy labels, is crucial for designing statistically robust solutions. Existing estimators for noise transition matrices, e.g., using either anchor points or clusterability, focus on computer vision tasks that are relatively easier to obtain high-quality representations. We observe that tasks with lower-quality features fail to meet the anchor-point or clusterability condition, due to the coexistence of both uninformative and informative representations. To handle this issue, we propose a generic and practical information-theoretic approach to down-weight the less informative parts of the lower-quality features. This improvement is crucial to identifying and estimating the label noise transition matrix. The salient technical challenge is to compute the relevant information-theoretical metrics using only noisy labels instead of clean ones. We prove that the celebrated $f$-mutual information measure can often preserve the order when calculated using noisy labels. We then build our transition matrix estimator using this distilled version of features. The necessity and effectiveness of the proposed method are also demonstrated by evaluating the estimation error on a varied set of tabular data and text classification tasks with lower-quality features. \rev{Code is available at \url{github.com/UCSC-REAL/BeyondImages}.}

\end{abstract}

% \vspace{-5pt}
\section{Introduction}\label{sec:intro}
% \vspace{-5pt}
When a feature is not properly annotated, the returned noisy label may differ from the ground-truth one \cite{wei2022learning}. One popular and useful statistical information about noisy labels is the \textit{label noise transition matrix} \cite{liu2022identifiability}, which characterizes the transition probability from the ground-truth label to a particular noisy label. The noise transition matrix has been demonstrated to be crucial in various tasks, e.g., learning noise-tolerant classifiers \cite{natarajan2013learning}, recovering unbiased estimates of fairness constraints \cite{lamy2019noise,wang2021fair}, identifying mislabeled data \cite{northcutt2021pervasive}, and aggregating multiple labels in crowdsourcing \cite{liu2015online}. These applications often postulate the true knowledge of the noise transition matrix, which is generally unattainable in real-world data. Problems may arise when the noise transition matrix, acquired through an estimation process, incurs misspecification. It has been shown that the blind application of such a noise transition matrix may in fact lead to higher errors \cite{liu2021can}. In consequence, it is important to estimate the label noise transition matrix accurately.

Most existing approaches to estimating the label noise transition matrix presuppose \emph{high-quality} features. For example, a series of research relies on finding the \textit{anchor points} \cite{scott2015rate,menon2015learning,patrini2017making,liu2015classification,xia2019anchor,yao2020dual,zhang2021learning,yang2021estimating}, defined as the instances that belong to a true label class almost surely. Without high-quality features, instances belonging to different true label classes may not be well-separated in any hidden space such that anchor points rarely exist. Likewise, estimating the noise transition matrix with confident learning \cite{northcutt2021confident} also requires high-quality features to estimate the confident points accurately.

Recent \textit{training-free} estimator requires $2$-Nearest-Neighbor ($2$-NN) clusterability \cite{zhu2021clusterability} to estimate the noise transition matrix, where one feature and its $2$-NN should have the same true label. This method inevitably depends on the quality of features since the likelihood that a data instance satisfies the clusterability condition will naturally decrease with lower-quality features.  

Despite the above approaches achieving huge success on image benchmark datasets, such as CIFAR-10 \cite{krizhevsky2009learning}, it remains questionable how these noise estimation approaches perform when it is hard to obtain high-quality features. In particular, for some tabular data with categorical features like UCI datasets \cite{dua2017uci}, popular feature learning or representation learning techniques seem unusable to improve the quality due to the sparseness of features. In more sophisticated natural language processing (NLP) tasks, such as text classification, some stop words (e.g., ``a/an'', ``the'', ``is'', and ``are'') are less informative but may be encoded in textual representations. For the learning-based methods, these less informative components may be overrated on noisy labels and harmful to finding anchor points or converging to the global optimum. For the training-free methods, these uninformative variables are likely to obscure the useful counterpart and result in mistakes in finding the nearest neighbors.

Unfortunately, we have observed consistent performance drops for the existing estimators, as shown in Figure~\ref{fig:failures}. We evaluate several baselines \cite{xia2019anchor,northcutt2021confident,zhu2021clusterability} and find the estimation errors of most approaches are around or larger than $0.1$ for binary classifications with lower-quality features. As a rough comparison, the estimation errors on CIFAR-10 reported by these methods are around $0.05$. Note the average noise rates of binary classification and a 10-class classification are within the range $[0,0.5)$ and $[0,0.9)$, respectively, showing the ``same'' error for binary classifications is effectively worse than a 10-class one, not to mention $0.1$ for binary versus an even lower $0.05$ for 10-class. Thus there is a non-negligible performance drop.
We defer more detailed observations and discussions to Section~\ref{sec:failures}.

We propose a generally practical information-theoretic approach to address the label noise matrix estimation for tasks with lower-quality features in response to the failures. Our approach builds on HOC \cite{zhu2021clusterability}, since we seek to avoid heavy task-specific hyperparameter fine-tuning and make it a generically applicable and light estimator. In particular, we divide the input features into several exclusively uncorrelated parts and down-weight the less informative parts by the $f$-mutual information between each part and the noisy labels. This operation allows us to improve the clusterability of features which proves to be a crucial property for identifying label noise transition matrix \cite{zhu2021clusterability,liu2022identifiability}. Our main contributions are:
\squishlist
\item Based on the $f$-mutual information, we propose a novel reweighting mechanism to firstly decouple features by projecting to its eigenspace and then down-weight the less-informative parts with only access to noisy labels. \rev{The mechanism intuitively disentangles features but it does not require training and can be applied efficiently.}
\item %
The salient technical challenge is to compute $f$-mutual information using only noisy labels instead of clean ones. We prove that calculating the total-variation-based mutual information with noisy labels preserves the same order as using clean labels (Theorem~\ref{thm:tvrobust})\rev{, and the} traditional mutual information preserves the order when the absolute gap between two noisy measurements is larger than a guaranteed threshold (Theorem~\ref{thm:kl}).
\item We empirically demonstrate \rev{that our approach helps return a more accurate estimate of the transition matrix and reduce the classification errors of downstream learning tasks on datasets with lower-quality or more sophisticated features, including UCI datasets with tabular data and text classification benchmarks.} %
\squishend

\subsection{Related Works}
The noise transition matrix $\mT$ is important in several communities \cite{han2020survey}. For example, it helps build noise-consistent classifiers when learning with noisy labels, such as loss correction \cite{natarajan2013learning,patrini2017making,wei2022aggregate}, loss reweighting \cite{liu2015classification}, and capturing the imbalance caused by heterogeneous noise rates \cite{zhu2020second}. It also helps tune hyperparameters for label smoothing \cite{lukasik2020does,wei2021understanding}, set thresholds for sample selection \cite{han2018co,wei2020combating,zhang2021alleviating} and detect label mistakes \cite{zhu2022detect,northcutt2021pervasive}. Additionally, $\mT$ contributes to evaluating \cite{awasthi2021evaluating} or improving \cite{lamy2019noise} the model fairness when the sensitive attribute is protected, or mitigating bias in treating different groups (e.g., racial, gender) when the label quality for different groups is different \cite{wang2021fair}.
It also has medical applications such as evaluating the physician variability \cite{mccormick2016probabilistic}.
All of the above applications require an accurate estimate of $\mT$. %

In addition to the $\mT$ estimators introduced in Section~\ref{sec:intro}, there are related works in crowdsourcing \cite{liu2012variational,zhang2014spectral,liu2015online} and peer prediction \cite{liu2017machine,liu2020surrogate}. However, these works require redundant noisy labels. For a general machine learning task, the datasets that have only one noisy label for each feature are more common. Compared with these approaches, the $2$-NN clusterability of HOC \cite{zhu2021clusterability} can be treated as a proxy of two redundant noisy labels, where a better proxy requires \rev{well-extracted features.
According to the analyses on the identifiability of the label noise transition matrix \cite{liu2022identifiability}, the disentangled and informative features are crucial.}
It has been demonstrated that mutual information helps select more informative features \cite{battiti1994using,estevez2009normalized,vergara2014review} with clean data. Recent work \cite{wei2020optimizing} finds that some $f$-mutual information metrics are robust to label noise, which makes the feature selection on noisy data promising.

\vspace{-2ex}
\section{Preliminaries}
\vspace{-3pt}
In this section, we first formally define the noise transition matrix $\mT$ and pose the problem in Section~\ref{sec:def_T}. As a complement to related works, we introduce more technical details of two lines of most relevant $\mT$ estimators in Section~\ref{sec:est_T} and numerically show the possible failures of these approaches in Section~\ref{sec:failures}.

\subsection{The Definition of Noise Transition Matrix}\label{sec:def_T}

\textbf{Clean/Noisy distribution~}
Consider a $K$-class classification task with a dataset $\widetilde D:=\{\vx_n,\tilde y_n\}_{n\in[N]}$, where $\vx_n$ is the feature, $\tilde y_n$ is the noisy label that may come from human annotations \cite{wei2022deep,wei2022learning,luo2020research}, sensors \cite{wang2021policy} and machine pseudo labels \cite{zhu2022the}, $N$ is the number of instances, $[N]:=\{1,2,\cdots,N\}$. Suppose $\vx$ is $d$-dimensional, i.e., $\vx = [x_1,\cdots,x_d]^\top$. We denote the $\mu$-th element by \textit{feature variable} $x_\mu$. Note the \textit{feature vector} $\vx$ is not necessary to be the raw input for some complicated tasks that require deep neural networks. In these tasks, the feature vector $\vx$ should be the output of some feature extractors \cite{devlin-etal-2019-bert,radford2021learning}. We will \emph{not} discuss methods to get appropriate feature extractors in this paper since the focus is on the data processing given $\widetilde D$. The clean label associated with the noisy label $\tilde y$ is denoted by $y$. Both clean labels and noisy labels are in the same label space, i.e., $y\in[K]$, $\tilde y\in[K]$, where $[K]:=\{1,2,\cdots,K\}$. The \textit{random variable} forms of the above realizations of features and labels are: {feature vector} $\vx \sim \mX:=[X_1,\cdots,X_d]^\top$, {feature variable} $x_\mu\sim X_\mu$, clean label $y\sim Y$, noisy label $\tilde y \sim \widetilde Y$. The (unobservable) clean dataset is denoted by $D$.

\textbf{Noise transition matrix $\mT$~}
The relationship between $(\mX,Y)$ and $(\mX,\widetilde Y)$ is denoted by a noise transition matrix $\mT(\mX)$, where each element $T_{ij}(\mX):= \PP(\widetilde Y=j|Y=i,\mX)$ stands for the probability of mislabeling a clean label $Y=i$ as the noisy label $\widetilde Y=j$ given feature $\mX$. A majority line of works assume the noise transition matrix is \textit{class-dependent} \cite{natarajan2013learning,liu2015classification,patrini2017making,liu2019peer}, i.e., $\mT(\mX) \equiv \mT$, implying the following independency:
$$
\PP(\widetilde Y=j|Y=i,\mX) = \PP(\widetilde Y=j|Y=i), \forall i,j\in[K], \forall \mX.
$$
\textbf{Problem Statement~}
We formally frame the problem of label noise transition matrix estimation. The learner can only access noisy examples $\widetilde D$ in this setting. The noisy label $\widetilde{Y}$ satisfies an underlying transition probability characterized by $\mT$. The goal is to minimize the estimation error calculated by the \textit{average total variation} of the true $\mT$ and the estimated $\hat \mT$ \cite{zhang2021learning}: 
\begin{equation}\label{eq:T_error}
    \textsf{Error}(\mT,\hat\mT) =  \sum_{i\in[K],j\in[K]} |T_{ij} - \hat T_{ij}|/(2K).
\end{equation}

\subsection{Existing Estimation Approaches}\label{sec:est_T}
There are mainly two categories of prior works on estimating the noise transition matrix $\mT$. One popular solution pipeline is to estimate $\mT$ with the confidence/predictions of the model \textit{trained on the noisy data distribution}. As an alternative to the learning-based methods, one recent work \cite{zhu2021clusterability} proposes a \textit{training-free} pipeline when \textit{clusterable} features are given. We introduce more details as follows.

\begin{definition}[Anchor Point \cite{liu2015classification,scott2015rate,xia2019anchor}]\label{def:anchor}
An instance $\vx$ is an anchor point for class-$i$ if $\PP(Y=i|\mX=\vx)$ is equal or close to $1$.
\end{definition}
\textbf{Using model predictions~}
In the learning-based methods, the anchor point, which is the instance that belongs to a particular class almost surely as defined in Definition~\ref{def:anchor}, plays a significant role. Particularly, if $\PP(Y=i|\mX=\vx)=1$ holds for some class $i$, we have
$
\PP(\widetilde Y=j|\mX=\vx) = \sum_{k\in[K]} T_{kj} \PP(Y=k,\mX=\vx) = T_{ij}.
$
Note the anchor point is defined on the clean data. If the features $\mX$ are of lower quality, the condition $\PP(Y=i|\mX=\vx)=1$ is hard to satisfy \cite{zhu2022detect}, not to mention finding anchor points accurately. Although recent advances relax the requirement of anchor points \cite{xia2019anchor,zhang2021learning,li2021provably}, they tend to design a specific learning pipeline with a particular regularizer or objective related to $\mT$, which is sensitive to hyperparameters and ultimately the quality of features.

\textbf{Using clusterability~}
\rev{Recent work HOC \cite{zhu2021clusterability} studies this problem from a novel data-centric perspective, which proposes a statistical solution based on clusterability without fitting the data distribution.}
The main idea is to utilize a set of \rev{High-Order Consensus (HOC)} information aggregated on the nearest neighbors' noisy labels and solve a series of equations for the noise transition matrix $\mT$ and clean label prior probability $\vp$.
The effectiveness of this approach relies on the $k$-NN ($k$-Nearest-Neighbor) label clusterability, which is defined as: 
\begin{definition}[$k$-NN label clusterability \cite{zhu2021clusterability}]\label{def:cluster}
A dataset $D$ satisfies $k$-NN label clusterability if $\forall n \in [N]$, the feature $\vx_n$ and its $k$-Nearest-Neighbor $\vx_{n_1}, \cdots, \vx_{n_k}$ belong to the same true label class.
\end{definition}
The distance between two features $\vx$ and $\vx'$ can be measured by $1-\textsf{Sim}(\vx,\vx')$, where $\textsf{Sim}(\vx,\vx')$ could be the cosine similarity.
It has been proved by \cite{zhu2021clusterability} that the $2$-NN label clusterability is sufficient for uniquely getting the true $\mT$. However, it is likely that the clusterability is not sufficiently satisfied for lower-quality features. 
\subsection{Failures on Lower-Quality Features}\label{sec:failures}
\begin{figure}[t]
    \begin{center}
    \centerline{
    \includegraphics[width=0.35\textwidth]{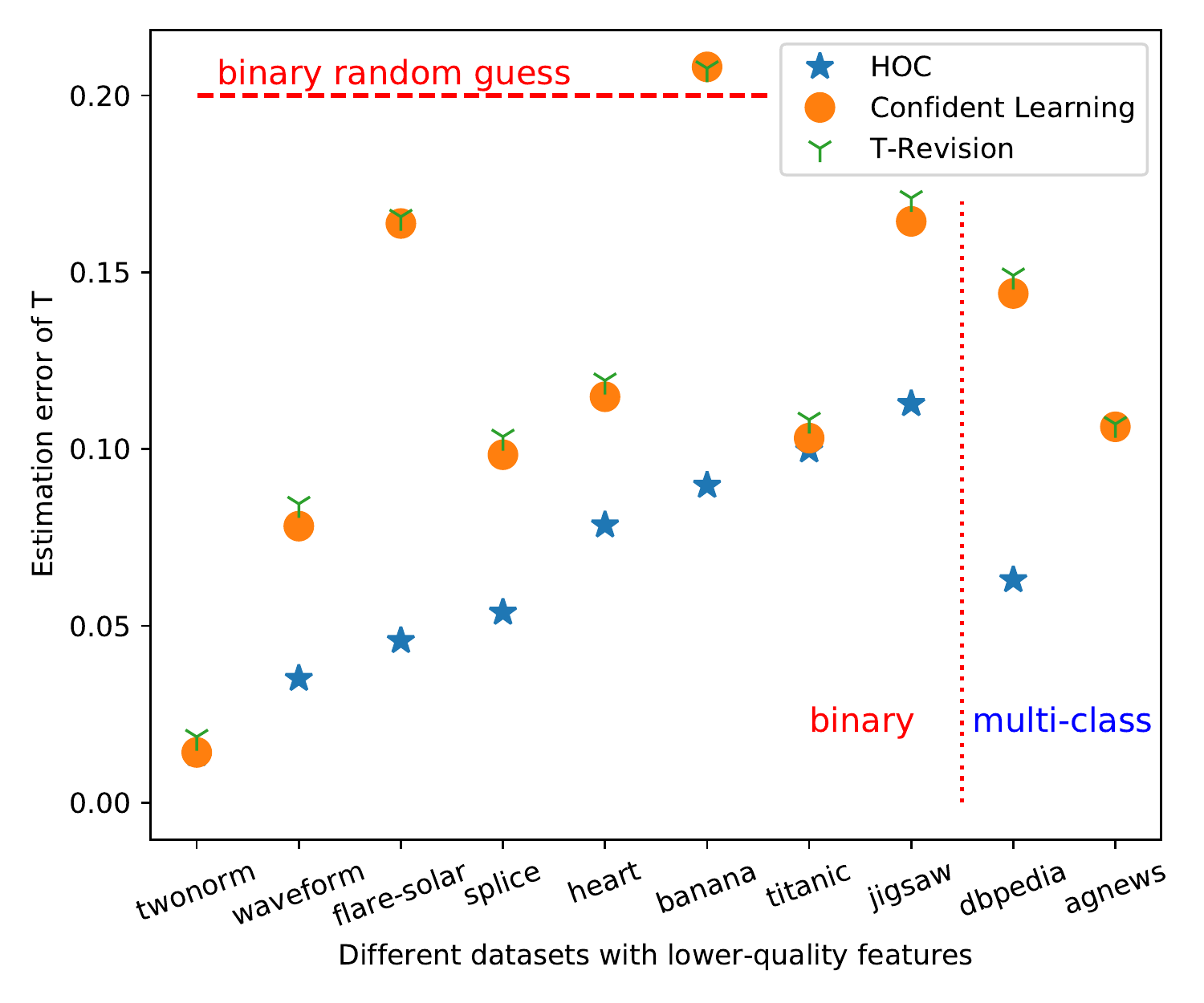}}
    \vspace{-8pt}
    \caption{Existing methods may suffer from failures. {\color{red}\textbf{Red}} horizontal dashed line shows the error of random guessing $\mT$ in binary classifications. Tasks on the left side of the dotted line are binary.}
    \vspace{-0.21in}
    \label{fig:failures}
    \end{center}
\vskip -0.2in %
\end{figure}
The above approaches to estimating $\mT$ are demonstrated to perform well on some image classification datasets, such as MNIST \cite{lecun1998gradient} and CIFAR \cite{krizhevsky2009learning}, \rev{which usually enjoy better representation learning tools \cite{chen2020simple,wang2022pico} that would return clusterable features, as compared to text sequence and tabular data. When facing other tasks,}  high-quality features may not always be available. In this subsection, we explore how these existing methods fare on other datasets, possibly with lower-quality features.

\textbf{Observations~}
In Figure~\ref{fig:failures}, we implement two learning-based methods built on anchor points (T-revision \cite{xia2019anchor}) or confident points (Confident Learning \cite{northcutt2021confident}), and one training-free method based on the clusterability (HOC \cite{zhu2021clusterability}). 
The average noise rate, i.e., $\sum_{i\in[K]}(1-T_{ii})/K$, is around $0.3$. Figure~\ref{fig:failures} shows the estimation error of these three methods on most of the datasets are around or larger than $0.1$, and some methods may even approximate to $0.2$. Note \textbf{an error of $0.2$ is excessive} for binary classification when the noise rate is $0.3$. For example, when $T_{11}=T_{22}=0.3$, a \textit{random guess} of $\mT$, i.e., $T_{ij} = 0.5, \forall i,j$, has an error of exactly $0.2$.
Recall an error of $0.05$ for binary classification is worse than the same error for a 10-class one as explained in Introduction.
Compared with an estimation error of $\approx 0.05$ on CIFAR-10 with a similar noise rate as reported in these baselines, Figure~\ref{fig:failures} shows a serious performance drop: an error of $0.05$ for $83\%$ of tests and an error of $0.1$ for $57\%$ of tests. 
Therefore, it is crucial to design an estimator which is also robust on datasets with lower-quality features.

\textbf{Discussions~}
Before diving into a concrete solution to lower-quality features, we discuss the advantages and disadvantages of both existing lines of work.
On one hand, the \textit{learning-based methods} \cite{northcutt2021confident,xia2019anchor} could take full advantage of deep neural networks. During supervised training, different parts of features are weighted differently, thus the informative parts could weigh more than the less informative ones. The optimal weight combinations could induce a model that accurately fits the data distribution, which further help estimate $\mT$. On the other hand, due to various factors such as the model capacity, the quality of features, the number of instances, and the setting of hyperparameters, the learning-based models are often infeasible to converge to the global optimum in practice. For example, with the existence of label noise,  deep neural networks (DNN) tend to be overconfident \cite{wei2022mitigating} and memorize wrong feature-label patterns \cite{cheng2021demystifying,liu2021importance,wei2021open}. When unintended memorization occurs, the weights for combining different parts will be non-optimal and some uninformative parts may be mis-specified with high weights.
Alternatively, the \textit{training-free method} \cite{zhu2021clusterability} will not be affected by the wrong memorization since it is a fully-statistical solution without any training procedures. Nevertheless, the problem of not employing training is also severe: blindly treating different parts of features equally important may cause failures. The above observations and discussions motivate us to find a solution that \textit{compromises between the learning-based and the training-free approaches}.

\section{An Information-Theoretic Approach}\label{sec:method}

We propose an information-theoretic approach to distinguish the importance of different features.
\rev{To avoid complicated hyperparameters tuning and make it a light tool for more general applications, the solution is built on HOC. See more detailed rationale in Appendix~\ref{appendix:rationale}. }

\rev{We now briefly introduce HOC \cite{zhu2021clusterability}. Algorithm~\ref{alg:hoc} summarizes the key steps, where the high-level idea is that, when $2$-NN label clusterability holds, the frequency of consensus patterns of the three grouped noisy labels $\tilde y_n,\tilde y_{n_1},\tilde y_{n_2}$ encodes $\mT$.\footnote{It is shown later in \cite{liu2022identifiability} that three noisy labels are necessary and sufficient to identify $\mT$.}
For instance, a triplet $\tilde y_n=0,\tilde y_{n_1}=0,\tilde y_{n_2}=1$ will add 1 to the count of the consensus pattern $(0,0,1)$.
With the estimated frequency of patterns $\{(\tilde y_n,\tilde y_{n_1},\tilde y_{n_2}), \forall n\}$, we can simply solve equations by gradient descents.
Therefore, in Algorithm~\ref{alg:hoc}, the $2$-NN label clusterability is critical, which depends heavily on calculating the distance or similarity between features.}

\begin{algorithm}[tb]
   \caption{Key Steps of HOC}
   \label{alg:hoc}
\begin{algorithmic}[1]
    \setalglineno{0}
   \STATE {\bfseries Input:} Noisy dataset: $\widetilde{D}=\{( \vx_n, \tilde y_n)\}_{n\in  [N]}$. 
    \\
    \algcom{// Find $2$-NN. $\textsf{Sim}(\bm x,\bm x') \rightarrow \textsf{Sim}_{\mW}(\bm z,\bm z')$ in our approach.}
    \STATE With $1-\textsf{Sim}(\bm x,\bm x')$ as the distance metric:\\
    $\{(\tilde y_n,\tilde y_{n_1},\tilde y_{n_2}), \forall n\}\leftarrow$ \texttt{Get2NN}$(\widetilde{D})$;
    \\
    \algcom{// Count first-, second, and third-order consensus patterns:}
    \STATE $({\hat{\vc}}^{[1]}$, ${\hat{\vc}}^{[2]}$,  ${\hat{\vc}}^{[3]})\leftarrow$  \texttt{CountFreq}($\{(\tilde y_n,\tilde y_{n_1},\tilde y_{n_2}), \forall n\}$)
    \\
    \algcom{// Solve equations}
    \STATE Find $\mT$ such that match the counts $({\hat{\vc}}^{[1]}$, ${\hat{\vc}}^{[2]}$,  ${\hat{\vc}}^{[3]})$.
\end{algorithmic}
\end{algorithm}
\begin{algorithm}[tb]
   \caption{Our Information-Theoretic Approach}
   \label{alg:Test}
\begin{algorithmic}[1]
    \setalglineno{0}
   \STATE {\bfseries Input:} Noisy dataset: $\widetilde{D}=\{( \vx_n, \tilde y_n)\}_{n\in  [N]}$. %
    \\
    \algcom{// Step 1:  Remove correlation (Section \ref{sec:remove_corr}) }
    \STATE Transform $\vx \sim \mX$ to $\vz \sim \mZ$ by Eqn.~(\ref{eq:projection});
    \\
    \algcom{// Step 2:  Estimate the weight matrix $\mW$ (Section \ref{sec:est_mi}) }
    \STATE With only noisy labels:\\
    \textit{Diagonal} elements: \quad ~$\hat W_{\mu\mu}= I_f(Z;\widetilde Y)$ by Eqn.~(\ref{eq:noisy_mi}) \\
    \textit{Off-diagonal} elements: $\hat W_{\mu\mu'} = 0, \forall \mu \ne \mu'$;
    \\
    \algcom{// Step 3: Estimate the noise transition matrix}
    \STATE Apply HOC \cite{zhu2021clusterability} with soft cosine similarity $\textsf{Sim}_{\mW}(\bm z,\bm z')$ defined in Eqn.~(\ref{eq:soft_cosine}).
    \STATE {\bfseries Output:} The estimated noise transition matrix $\hat\mT$.
\end{algorithmic}
\end{algorithm}

\rev{\textbf{Overview of our approach:~}
The main idea is to decouple features (Step 1) and down-weight the less informative parts (Step 2) when measuring the distances by HOC, which is summarized in Algorithm~\ref{alg:Test}.
Firstly, we motivate the necessity of down-weighting less informative parts in Section~\ref{sec:motivation}. Their weights are controlled by a matrix $\mW$, which is absorbed into the calculation of soft cosine similarity (Definition~\ref{def:softcosine}). A tractable proxy of $\mW$ is introduced in Section~\ref{sec:proxy_W} and detailed in the remainder of this section.}

\subsection{Vanilla Similarity Measures Are Not Sufficient}\label{sec:motivation}

Our following analyses focus on the cosine similarity \rev{as originally implemented by HOC \cite{zhu2021clusterability}}. The larger cosine similarity implies the smaller distances of features. We first analyze possible problems in evaluating $k$-NN with the vanilla (hard) cosine similarity.

Consider the cosine similarity of two feature vectors $\vx$ and $\vx'$, which is denoted by
$\textsf{Sim}(\vx,\vx') = \frac{\vx^\top \vx'}{\|\vx\|_2 \|\vx'\|_2},$
where $\|\cdot\|_2$ denotes the vector $\ell_2$ norm. 
The above measure inherently assumes different elements of $\vx$ are 1) \emph{equally important} and 2) \emph{uncorrelated} to each other, thus may underestimate the true similarity between imperfect feature vectors. 
To capture more information, we incorporate a change-of-basis matrix $\sqrt{\mW}$ to obtain a soft cosine measure defined as follows.
\begin{definition}[Soft cosine similarity \cite{Sidorov2014SoftSA}]\label{def:softcosine}
\begin{equation}\label{eq:soft_cosine}
    \textsf{Sim}_{\mW}(\vx,\vx') = \frac{(\sqrt{\mW}\vx)^\top (\sqrt{\mW}\vx')}{\|\sqrt{\mW}\vx\|_2 \|\sqrt{\mW}\vx'\|_2}.
\end{equation}
\end{definition}

Hereby, the symmetric matrix $\mW$ encodes the pairwise similarity between features. Note that the soft cosine similarity measure $\textsf{Sim}_{\mW}(\vx,\vx')$ recovers the (hard) cosine similarity when $\mW = \mI$, where $\mI$ denotes a $K\times K$ identity matrix.
In practice, the true and unknown $\mW$ may be very different from $\mI$. Thus simply letting $\mW=\mI$ may cause severe problems in using clusterability. 
For example, when $K=2$, consider three instances $(\vx_1,y_1) = ([1,0,1]^\top, 1)$, $(\vx_2,y_2) = ([0,1,0]^\top,2)$, and $(\vx_3,y_3) = ([0.8,1,0.7]^\top,y)$.
Based on $1$-NN label clusterability, we infer label $y$ following the rule below:
\[
y=\begin{cases}
1 \qquad  \textit{~if~} \quad\textsf{Sim}(\vx_1,\vx_3) > \textsf{Sim}(\vx_2,\vx_3);\\
2 \qquad  \textit{~if~} \quad\textsf{Sim}(\vx_1,\vx_3) \le \textsf{Sim}(\vx_2,\vx_3).
\end{cases}
\]
Consider the following three $\mW$s: $\mW_1 = \mI$,
{\small\[
\mW_2 = \begin{pmatrix}
1.0 & 0.0 & 0.0 \\
0.0 & 1.0 & 0.0\\
0.0 & 0.0 & \textbf{0.1}
\end{pmatrix}, 
~~ \mW_3 =\begin{pmatrix}
1.0 & -0.2 & -0.5 \\
-0.2 & 1.0 & 0.5\\
-0.5 & 0.5 & 1.0 
\end{pmatrix}.
\]}

\textbf{Example 1: $\hspace{-1pt}\mW_1\hspace{-1pt}$  (uncorrelated and equally important)}~\\
The following hard cosine similarity shows:\\
$\textsf{Sim}_{\mW_1}(\vx_1,\vx_3) \hspace{-2pt}\approx\hspace{-2pt} 0.73> \textsf{Sim}_{\mW_1}(\vx_2,\vx_3) \hspace{-2pt} \approx\hspace{-2pt}  0.69 \hspace{-1pt}\Rightarrow\hspace{-1pt} y \hspace{-1pt}=\hspace{-1pt} 1.$

\textbf{Example 2: $\mW_2$ (not equally important)}~\\
The following soft cosine similarity shows: \\
$\textsf{Sim}_{\mW_2}(\vx_1,\vx_3) \hspace{-2pt}\approx\hspace{-2pt} 0.64 < \textsf{Sim}_{\mW_2}(\vx_2,\vx_3) \hspace{-2pt}\approx\hspace{-2pt} 0.77 \hspace{-2pt}\Rightarrow y \hspace{-2pt}= 2,$
which is \textit{different} from the inferred $y$ in Example 1. If $\mW_2$ defines the true clusterability, this example shows that simply using $\mW=\mI$ fails to capture the diagonal values of $\mW$ (the weights of $x_\mu$, $\mu\in[d]$) and violates the clusterability.

\textbf{Example 3: $\mW_3$ (correlated)}~\\
The following soft cosine similarity shows:\\
$\textsf{Sim}_{\mW_3}(\vx_1,\vx_3) \hspace{-2pt}\approx\hspace{-2pt} 0.75 < \textsf{Sim}_{\mW_3}(\vx_2,\vx_3) \hspace{-2pt}\approx\hspace{-2pt} 0.85 \hspace{-2pt}\Rightarrow \hspace{-2pt} y=2,$
which is \textit{different} from the inferred $y$ in Example 1. If $\mW_3$ is true, this example shows that simply using $\mW=\mI$ fails to capture the off-diagonal values of $\mW$ (the correlations between $x_\mu$) and violates the clusterability.

The above three examples exemplify that the less informative parts of features may damage the clusterability, where the definition of each ``part'' depends on both the diagonal elements and off-diagonal elements of $\mW$. We propose an information-theoretic approach to construct a proxy of $\mW$.

\subsection{Proxy of \texorpdfstring{$\mW$}{}}\label{sec:proxy_W}

Noting $\mW$ is a square matrix of order $d$, it requires $\mathcal{O}(d^2)$ operations to estimate all elements. Each operation incurs an estimation error, and the accumulated errors may not be bounded. We propose to find a proxy of $\mW$.
Intuitively, we expect the following properties \cite{battiti1994using}:
\squishlist 
\item \textbf{Symmetric}: ~$\forall \mu, \nu \in [d], W_{\mu\nu} = W_{\nu \mu}$. 
\item \textbf{Information monotone}: For every two feature variables $X_\mu, X_\nu$, if $X_\mu$ is less informative with respect to $Y$ than $X_\nu$, then $X_\mu$ will be less important than $X_\nu$, i.e.,
$
    I_f(X_\mu;Y) \leq I_f(X_\nu;Y) \Rightarrow  W_{\mu\mu} \leq W_{\nu\nu}, \mu,\nu\in[d],
$
where $I_f(X_\mu;Y)$ measures the \rev{$f$-MI ($f$-Mutual Information)} \cite{csiszar1967information} between $X_\mu$ and $Y$. 
\item \textbf{Correlation monotone}: Given two feature variables $X_\mu, X_\nu$, for any other feature variable $X_{\nu'}$ ($X_\mu,X_\nu,X_{\nu'}$ are equally informative), if $X_\mu$ is less correlated to $X_\nu$ than $X_{\nu'}$, $X_\nu$ will have a lower weight than $X_{\nu'}$ in measuring the similarity with $X_\nu$:
\[
    \rho(X_\mu,X_\nu) \leq \rho(X_{\mu},X_{\nu'}) \Rightarrow  W_{\mu\nu} \leq W_{\mu\nu'},
\]
where $\rho(X_\mu,X_\nu)$ is the correlation between random variables $X_\mu$ and $X_\nu$.
\squishend

\begin{figure}[t]
    \centering
    \includegraphics[width=0.4\textwidth]{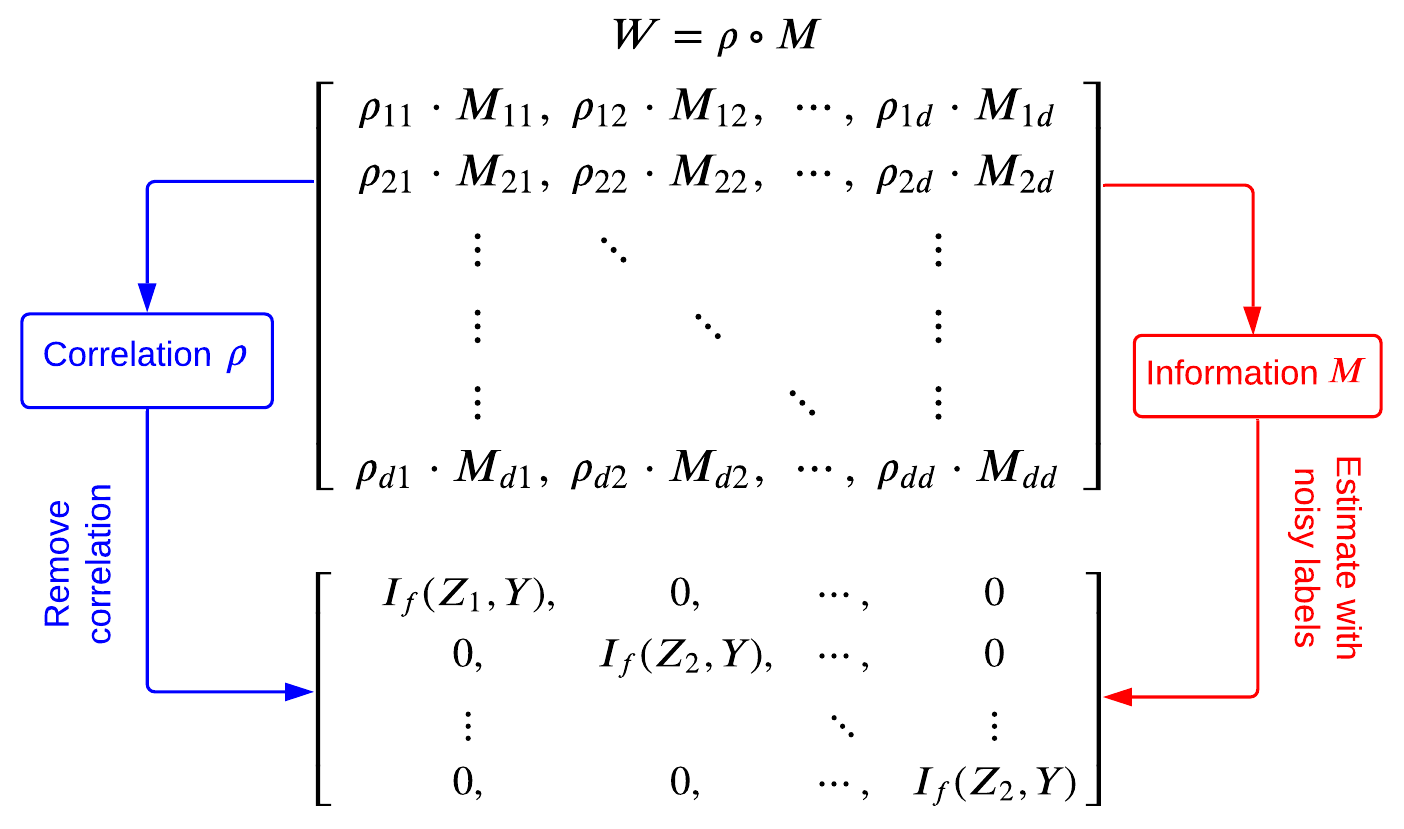}
    \vspace{-0.1in}
    \caption{Illustration of the proxy of $\mW$.
    }
    \vspace{-0.2in}
    \label{fig:est_W}
\end{figure}

The above properties suggest us to decompose the elements of matrix $\mW$ as the product of the informativeness w.r.t $Y$ and the correlation between features as illustrated in Figure~\ref{fig:est_W}. In another word, we can construct the matrix $\mW = \bm\rho \circ \mM$, where $\circ$ denotes the Hadamard product of two matrices, $\bm \rho$ is the correlation matrix with $\rho_{\mu\nu} = \rho(X_\mu,X_\nu)$, $\mM$ includes the $f$-MI with $M_{\mu\nu} = \sqrt{I_f(X_\mu,Y)I_f(X_\nu,Y)}$.
Directly estimating $\bm\rho$ might not be computation-efficient. If we can transform $\mX$ to make $\bm\rho = \bm I$, the off-diagonal entries of $\mW$ will become $0$ thus no longer need to be estimated. This observation motivates us to firstly transform $\mX$ to a non-correlated form to achieve $\bm \rho=\mI$ (Section~\ref{sec:remove_corr}), then estimate only the diagonal elements by $f$-MI (Section~\ref{sec:est_mi}).

\subsection{Remove correlation}\label{sec:remove_corr}
For ease of notations, we require that $\mX$ is zero-mean, i.e., $\E[\mX] = \bm 0$, where $\bm 0$ is a $d\times 1$ column vector with all elements being $0$.
\rev{Inspired by the principle component analysis (PCA) \cite{wold1987principal},
we} adopt $\bm\Lambda^{-1/2}\mP^\top$ as the matrix to remove the correlation between $x_\mu, x_\nu, \mu\ne \nu$, where $\mP \bm \Lambda \mP^\top = \E[\mX \mX^\top]$, $\mP$ is an orthogonal matrix whose columns are eigenvectors of $\E[\mX \mX^\top]$, $\bm\Lambda$ is a diagonal matrix where diagonal values are eigen values and off-diagonal values are $0$. Let 
\begin{equation}\label{eq:projection}
    \mZ = \bm\Lambda^{-1/2}\mP^\top \mX = [Z_1,\cdots,Z_d]^\top,
\end{equation}
where $\E[\mZ] = \bm\Lambda^{-1/2}\mP^\top \E[\mX]  = \bm 0$.
We have
\[
\E[\mZ \mZ^\top] = \bm\Lambda^{-1/2}\mP^\top \E[\mX \mX^\top] \mP \bm\Lambda^{-1/2} = \mI.
\]
Therefore, after transforming $\mX$ to $\mZ$ by $\bm\Lambda^{-1/2}\mP^\top$, we can split $\mX$ into $d$ uncorrelated parts such that $\E[Z_\mu Z_\nu]$ satisfies
\[
\E[Z_\mu Z_\nu] = \E[(\mZ\mZ^\top)_{\mu,\nu}] = \begin{cases}
0 \qquad \textit{~if~} \mu \ne \nu, \\
1 \qquad \textit{~if~} \mu = \nu.
\end{cases}
\]
Noting (zero-mean) $\rho(Z_\mu,Z_\nu) = \frac{\E[Z_\mu Z_\nu]}{\sqrt{\E[Z_\mu^2]}\sqrt{\E[Z_\nu^2]}},$
we know $\bm \rho = \mI$ after this transformation.

\subsection{Estimate $I_f(Z;Y)$ With Only Noisy Labels}\label{sec:est_mi}
After transforming $\mX$ to $\mZ$, the off-diagonal elements of the proxy of $\mW$ are expected to be $0$. Thus we only need to estimate the diagonal elements by $W_{\mu\mu}:=I_f(Z_\mu;Y)$.

Define an $f$-MI metric with $f$-divergence $I_f(Z;Y):= D_f(Z \oplus Y;Z\otimes Y)$, where $D_f(P||Q)$ measures the $f$-divergence between two distributions $P$ and $Q$ with probability density function $p$ and $q$:
\[
D_f(P||Q) = \int_{v\in\mathcal V} q(v) f\left( \frac{p(v)}{q(v)} \right) ~dv.
\]
Note the variable $v$ in the domain $\mathcal V$ is a realization of random variable $V=(Z,Y)$, which follows either distribution $P$ or distribution $Q$ depending on where it is used.
In our setting, $P$ and $Q$ are the joint distribution $P:=Z \oplus Y = \PP(Z=z, Y=y)$ and the marginal product distribution $Q:=Z\otimes Y = \PP(Z=z) \cdot \PP(Y=y)$, respectively. There are lots of choices of $f$.
Specially, when $f(v) = v\log v$, the $f$-divergence becomes the celebrated KL divergence and $I_f$ is exactly the mutual information.

Alternatively, we can calculate the $f$-divergence using the variational form \cite{nowozin2016f,wei2020optimizing}.
Denote the variational difference between $P$ and $Q$ by
\[
\textsf{VD}_f(g) = \E_{V\sim P}[g(V)] - \E_{V\sim Q}[f^*(g(V))], \forall g,
\]
where $g:\mathcal{V}\rightarrow \text{domain}(f^*)$ is a variational function, and $f^*$ is the conjugate function of $f(\cdot)$. 
We instantiate some prominent variational-conjugate $(g^*,f^*)$ pairs in Appendix~\ref{appendix:f-div} (Table~\ref{table:f_div-full}).
The $f$-MI is calculated by
\[
I_f(Z;Y) = D_f(P||Q) = \textsf{VD}_f(g^*) = \sup_{g} ~ \textsf{VD}_f(g).
\]
However, the above calculation, built on the clean distributions, will be intractable when we can only access the noisy data distribution.
Let $\widetilde P := Z \oplus \widetilde Y$ and $\widetilde Q := Z \otimes \widetilde Y$.
One tractable approach is to calculate $D_f(\widetilde P||\widetilde Q)$ following
\begin{equation}\label{eq:noisy_mi}
    I_f(Z;\widetilde Y) = D_f(\widetilde P||\widetilde Q) = \widetilde{\textsf{VD}}_f(\tilde g^*) = \sup_{g} ~ \widetilde{\textsf{VD}}_f(g),
\end{equation}
where
$
\widetilde{\textsf{VD}}_f(g) = \E_{\widetilde V\sim \widetilde P}[g(\widetilde V)] - \E_{\widetilde V\sim \widetilde Q}[f^*(g(\widetilde V))], \forall g.
$
Generally, there will be a gap between our calculated $I_f(Z;\widetilde Y)$ and the real $I_f(Z;Y)$. We defer detailed analyses of the gap to Section~\ref{sec:theorem}.

\vspace{-3pt}
\section{Theoretical Guarantees}\label{sec:theorem}
\vspace{-3pt}
\rev{
The quality of our $\mT$ estimator relies on the following steps:
\begin{itemize}[itemsep=0pt,topsep=-3pt,parsep=0pt, leftmargin=15pt]
    \item [(a)] Noise-resistant estimates of $f$-MI using noisy labels; 
    \item [(b)] Accurate estimates of clean $f$-MI;
    \item [(c)] Down-weighting less informative features with $f$-MI;
    \item [(d)] Robust distance/similarity calculation;
    \item [(e)] Satisfying clusterability (Definition~\ref{def:cluster});
    \item [(f)] Accurate estimates of the noise transition matrix $\mT$. 
\end{itemize}
The most critical step in the above chain is (a)$\rightarrow$(b), which
is the key ingredient to Step (c). This step will be the focus of the theoretical results in this section. Steps (c)$\rightarrow$(e) are explained in Section~\ref{sec:method}. Steps (e)$\rightarrow$(f) is guaranteed by HOC \cite{zhu2021clusterability}.
}

Based on our intuition for constructing the proxy of $\mW$ that the less informative parts should be assigned with lower weights, the order between two parts with different informativeness is crucial. Thus in this section, we study whether the noisy $f$-MI calculated using noisy labels, i.e., $I_f(Z;\widetilde Y)$, preserves the order of the clean $f$-MI $I_f(Z;Y)$.
\rev{Note the order-preservation property distinguishes from the robustness of $f$-MI between optimal classifier prediction $h^*(X)$ and noisy label $\widetilde Y$ by \citet{wei2020optimizing} since 1) $Z$ does not have class-specific meaning; 2) $Z$ is not optimizable and its ranking is concerned, while $h^*(X)$ is only a special (optimal) case of $Z$.}
All proofs are deferred to Appendix~\ref{appendix:thm}.
We define $\epsilon$-order-preserving as follows.
\begin{definition}[$\epsilon$-Order-Preserving Under Label Noise]
$I_f(Z;\widetilde Y)$ is called $\epsilon$-order-preserving under label noise if $ \forall \mu \in [d], \nu \in [d]$, given
$
|I_f(Z_\mu;\widetilde Y) - I_f(Z_\nu;\widetilde Y)| > \epsilon,
$
we have $$\sign [I_f(Z_\mu;\widetilde Y) - I_f(Z_\nu;\widetilde Y) ]     =  \sign [I_f(Z_\mu; Y) - I_f(Z_\nu; Y) ].$$
\end{definition}
The smaller $\epsilon$ is, the stricter the requirement is. The following analyses focus on the binary classification. Define $e_1 := \PP(\widetilde Y=2|Y=1)$, $e_2 := \PP(\widetilde Y=1|Y=2)$.
To show an $f$-MI metric is $\epsilon$-order-preserving under label noise, we need to study how $\widetilde{\textsf{VD}}_f(\tilde g^*)$ differs from the order of  ${\textsf{VD}}_f(g^*)$. 

\subsection{Total-Variation is $0$-Order-Preserving}
When $f(v) = \frac{1}{2}|v-1|$, we get the Total-Variation (TV). To analyze the order-preserving property of TV, we first build the relationship between $\widetilde{\textsf{VD}}_f(g)$ and $\widetilde{\textsf{VD}}_f(g)$, $\forall g$ by the following lemma:
\begin{lemma}[Linear relationship \cite{wei2020optimizing}]\label{lem:linear}
\[
\widetilde{\textsf{VD}}_{\text{TV}}(g) = (1-e_1 - e_2) {\textsf{VD}}_{\text{TV}}(g), \forall g.
\]
\end{lemma}
Lemma~\ref{lem:linear} shows that there is a linear relationship between $\widetilde{\textsf{VD}}_{\text{TV}}(g)$ and ${\textsf{VD}}_{\text{TV}}(g)$. The constant only depends on the noise rates.
With this lemma, we only need to study the difference between ${\textsf{VD}}_{\text{TV}}(\tilde g^*)$ and ${\textsf{VD}}_{\text{TV}}(g^*)$, which is summarized in the following lemma:
\begin{lemma}\label{lem:tv_gap}
When $e_1+e_2<1$,
${\textsf{VD}}_{\text{TV}}(\tilde g^*) = {\textsf{VD}}_{\text{TV}}(g^*)$.
\end{lemma}
Note the condition $e_1+e_2<1$ indicates the label noise is not too large to be dominant \cite{liu2019peer,natarajan2013learning,liu2017machine}.
With Lemma~\ref{lem:linear} and Lemma~\ref{lem:tv_gap}, we can conclude that TV is $0$-order-preserving.

\begin{theorem}\label{thm:tvrobust}
When $e_1+e_2<1$, total-variation is $0$-order-preserving under class-dependent label noise.
\end{theorem}
Theorem~\ref{thm:tvrobust} shows the total-variation-based mutual information under class-dependent label noise preserves the order of the original clean results.
\begin{figure}[!t]
    \begin{center}
    \centerline{\includegraphics[width=0.48\textwidth]{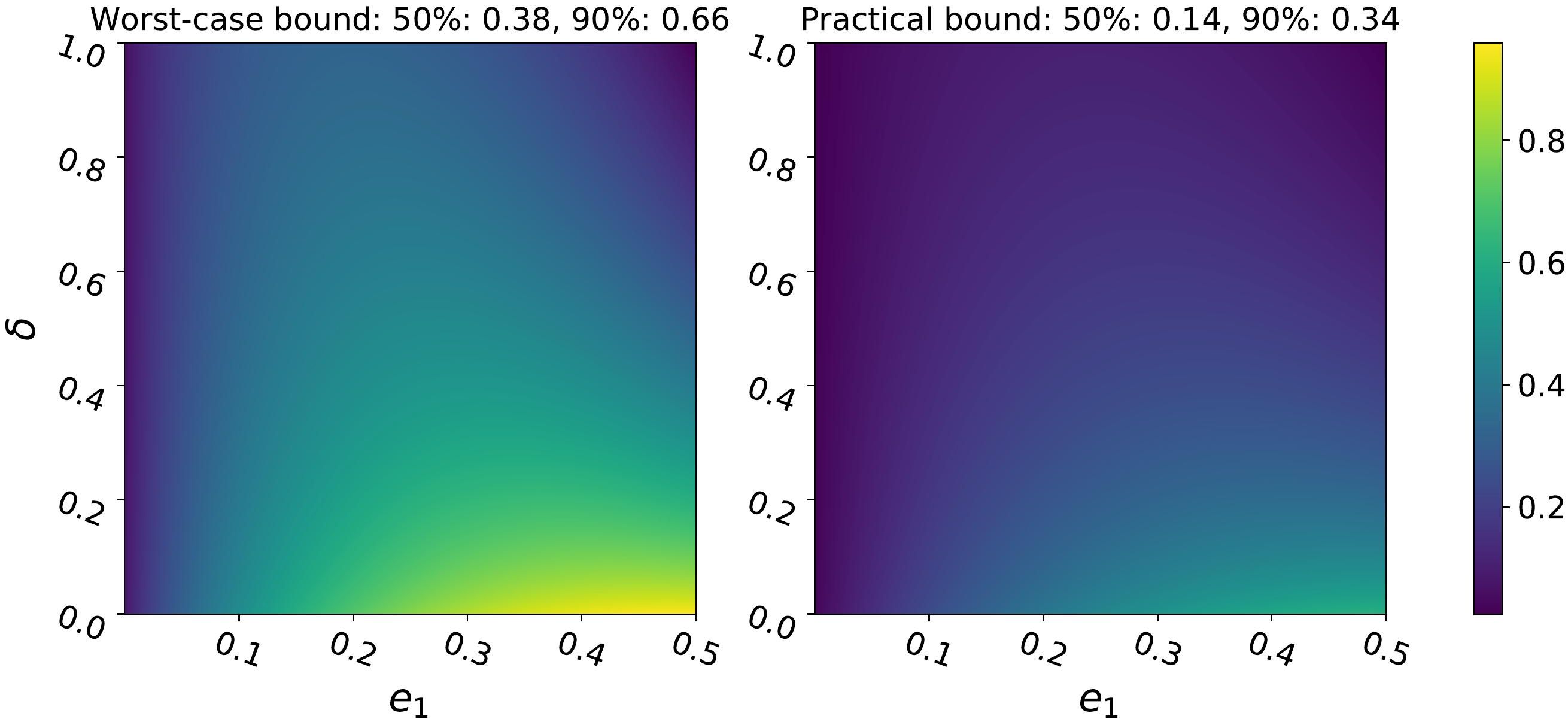}}
    \vspace{-0.1in}
    \caption{Illustration of the worst-case bound and a more practical bound for $\epsilon$ with different $e_1, \delta := e_2/e_1$. Color indicates the value of $\epsilon$. The median (50\%) and the 90-th percentile of $\epsilon$ are shown in the title of each plot. $\log_2(x)$ is applied for calculating mutual information.}
    \label{fig:gap}
    \vspace{-0.2in}
    \end{center}
\vskip -0.2in %
\end{figure}

\subsection{KL Divergence is $\epsilon$-Order-Preserving}
Unfortunately, Lemma~\ref{lem:linear} and Lemma~\ref{lem:tv_gap} do not hold for KL divergence. 
\rev{Recall the $f$-MI is exactly the standard mutual information when KL divergence is adopted.
Denote by $H(\cdot)$ the entropy.
We can start from the definition of mutual information $I(Z,\widetilde Y) = H(Z) + H(\widetilde Y) - H(Z,\widetilde Y)$ and decouple the effect of noisy labels as:
\begin{align*}
    I(Z,\widetilde Y)= &
(1-e_1-e_2) \cdot  [I(Z,Y) - H(Y)] + {H(\widetilde Y)} \\
& + \int_z \Delta_{\textsf{Bias}}(\beta_z, e_1, e_2)  ~dz, 
\end{align*}
where $\beta_z$ is a function of $z$, and $\Delta_{\textsf{Bias}}(\beta_z, e_1, e_2)$ is the bias caused by label noise specified in Eqn.~(\ref{eqn:delta_bias_beta}).
We then find the lower and upper bounds for $\Delta_{\textsf{Bias}}(\beta_z, e_1, e_2)$ and summarize the result as follows:}
\begin{theorem}\label{thm:kl}
Assume $e_1 =\delta e_2$, $\delta\in[0,1]$ and $e_1+e_2<1$. KL divergence (mutual information) is $\epsilon$-order-preserving under class-dependent label noise, where
\[
\epsilon =  e_1 \left[ \delta\log \delta - (1+\delta)\log (1+\delta)     \right]  + H(e_1),
\]
and $H(e_1):=-e_1\log e_1 - (1-e_1)\log(1-e_1)$. 
\end{theorem}

For the symmetric label noise, we have:
\begin{corollary}
When $e_1 = e_2<0.5$, KL divergence (mutual information) is $[H(e_1)-2e_1\log2]$-order-preserving under class-dependent label noise.
\end{corollary}
We evaluate the bound in the following remark.
\begin{remark}
The value of $\epsilon$ is illustrated in Figure~\ref{fig:gap}, where the left figure shows the worst-case bound in Theorem~\ref{thm:kl}, and the right figure shows the case when $\frac{\PP(Y=1|Z=z)}{\PP(Y=2|Z=z)} \in [\frac{1}{5},5]$. This is a more practical case for lower-quality features since a single feature variable cannot infer the clean label with high confidence. Noting the $\log_2$-based mutual information is within range $[0,1]$ and $\epsilon \le 0.34$ happens %
in $90\%$ of the cases, it is reasonable to believe the mutual information has a good $\epsilon$-order-preserving ability. 
\end{remark}

\vspace{-0.1in}
\section{Evaluations}

\begin{table*}[thb]
\caption{The estimation error ($\times 100$) on tabular benchmarks. Feature dimension: $d$. Number of clean instances in class-1 (or class-2): $N_{1}$ (or $N_2$). Noise Rate: {\sc Low: $e_1=0.2,e_2=0.1$. Medium: $e_1=0.4,e_2=0.2$. High: $e_1=0.4,e_2=0.4$.} Top-2 of each row are \textbf{bold}.}
\label{tab:exp1}
\vskip 0.1in %
\begin{center}
    \begin{small}
    \begin{sc}
    \resizebox{0.8\linewidth}{!}{
\begin{tabular}{c c|c c c c c c c c}
\toprule
\multicolumn{2}{c|}{Tabular Datasets}  & \multicolumn{7}{c}{Method}            \\ 
$(d,[N_1, N_2])$ & Noise Rate  & T-Rev  & CL  & HOC &  Ours-$\mX$-KL & Ours-$\mX$-TV & Ours-$\mA$-KL & Ours-$\mA$-TV  \\ 
\midrule\midrule
               & Low  & 9.25 & \textbf{8.00} & 9.82 & \textbf{8.09} & 8.70 & 8.88 & 9.18 \\
Heart        & Medium  &11.94&	11.48&	7.85&		9.55&	\textbf{1.48}&	\textbf{3.98}&	5.51\\ 
(23, [138, 165]) & High  & 6.74&	6.54&	4.71&		9.78&	14.91&	\textbf{1.33}&	\textbf{4.21} \\ 
\midrule
               & Low  & 30.89 & 30.53 & 14.96 & 10.74 & 11.84 & \textbf{10.57} & \textbf{9.26} \\ 
Banana        & Medium  & 20.77&	20.81&	8.97&		\textbf{4.90}&	\textbf{3.98}&	6.41&	6.97 \\ 
(2, [2924, 2376]) & High  &\textbf{ 6.71}&	7.58&	\textbf{4.78}&		12.11&	8.89&	9.78&	11.26 \\ 
\midrule
               & Low  & 21.40 &	20.62 &	11.24 &	\textbf{10.83} & \textbf{9.96} & 11.05 & 11.60
\\ 
Titanic        & Medium  & 10.83&	10.31&	9.97&		9.82&	\textbf{9.65}&	\textbf{9.61}&	9.75 \\ 
(3, [1490, 711]) & High  & 6.93&	6.75&	1.94&		1.97&	1.97&	\textbf{1.89}&	\textbf{1.92} \\ 
\midrule
               & Low  & 10.63 &	9.32&	7.32&	\textbf{2.29}&	\textbf{1.87}&	3.00&	3.65\\ 
Splice        & Medium  & 10.35&	9.84&	5.38	&	2.21&	3.90&	\textbf{1.26}&	\textbf{2.15} \\ 
(240, [1648, 1527]) & High  & 7.86&	7.74&	17.43&		\textbf{4.26}&	\textbf{2.94}&	4.41&	5.81&
 \\ 
 \midrule
               & Low  & 1.79 &	1.63&	2.12&		2.34&	2.80&	\textbf{0.54}&	\textbf{0.65} \\ 
Twonorm        & Medium & 1.86&	1.42&	\textbf{1.38}&		1.42&	\textbf{1.30}&	2.14&	1.73\\ 
(20, [3697, 3703]) & High  & 1.67&	\textbf{1.22}&	5.18&		5.88&	3.47&	\textbf{1.46}&	2.12 \\ 
\midrule
               & Low  & 10.59 &	10.89&	7.93	&	6.68&	6.78&	\textbf{5.67}&	\textbf{5.79} \\ 
Waveform        & Medium  & 8.45&	7.82&	3.51&		\textbf{2.50}&	3.09&	\textbf{2.34}&	2.72 \\ 
(21, [3353, 1647]) & High  & 5.04&	4.76&	3.84&		2.72&	\textbf{1.71}&	\textbf{2.29}&	5.42 \\ 
\midrule
               & Low  & 19.28&	18.43&	15.24	&	\textbf{14.60}&	14.84&	15.63&	\textbf{14.71} \\ 
Flare-solar        & Medium  & 16.57&	16.38&	\textbf{4.58}&		\textbf{4.39}&	5.05&	4.64&	4.82 \\ 
(31, [477, 589]) & High  & 8.35&	8.25&	4.71&		4.47&	4.74&	\textbf{3.87}&	\textbf{3.30} \\
\bottomrule
\end{tabular}}
\vspace{-0.1in}
\end{sc}
    \end{small}
\end{center}
\vskip -0.1in %
\end{table*}

\begin{table*}[thb]
\caption{The estimation error ($\times 100$) on natural language benchmarks. Feature dimension: $d$. Number of clean instances in class-$k$: $N_{k}$. {\sc [30k $\times$ 4]}: $N_1=N_2=N_3=N_4=$30k. Average noise rate follows $e={1}/{(1+r/\sqrt{K-1})}$. {\sc Low: $r=8$. Medium: $r=4$. High: $r=1.5$.} Top-2 of each row are \textbf{bold}.}
\label{tab:exp2}
\vskip 0.1in %
\begin{center}
    \begin{small}
    \begin{sc}
\resizebox{0.8\linewidth}{!}{
\begin{tabular}{c c|c c c c c c c c}
\toprule
\multicolumn{2}{c|}{Text Datasets}  & \multicolumn{7}{c}{Method}            \\ 
$(d,[N_1, \cdots, N_K])$ & Noise Rate  & T-Rev  & CL  & HOC &  Ours-$\mX$-KL & Ours-$\mX$-TV & Ours-$\mA$-KL & Ours-$\mA$-TV  \\ \midrule \midrule
               & Low  & 10.38&	11.41&	13.32	&	12.65&	12.75&	\textbf{8.36}&	\textbf{8.35}  \\ 
AG's news (BERT)       & Medium  & 10.71&	10.63&	10.62	&	10.13&	10.45&	\textbf{6.44}&	\textbf{6.52 } \\ 
(768, [30k $\times$ 4]) & High  & 13.97 &	13.82&	6.80&		6.83&	6.69&	\textbf{4.54}&	\textbf{4.19}  \\  \midrule
               & Low  & 6.80&	5.31&	7.57&		6.76&	6.94&	\textbf{2.52}& \textbf{2.52}  \\ 
DBpedia (BERT)       & Medium  & 14.91&	14.40&	6.30&		5.66&	5.78&	\textbf{2.33}&	\textbf{2.28} \\ 
(768, [40k $\times$ 14]) & High  & 24.23&	23.28&	6.00&		5.18&	5.22&	\textbf{2.42}&	\textbf{2.43} \\  \midrule
               & Low  & 38.49&	38.75&	40.87&		40.71&	40.58&	\textbf{37.37}&	\textbf{37.19 }\\ 
Yelp-5 (BERT)         & Medium  & 35.46&	36.05&	33.63&		34.23&	33.88&	\textbf{31.79}&	\textbf{31.94}  \\ 
(768, [130k $\times$ 5]) & High  & 21.20&	20.88&	19.09&		18.56&	20.13&	\textbf{18.11}&	\textbf{18.06} \\  \midrule
               & Low  & 20.92&	20.17&	14.25&		14.07&	14.24&	\textbf{9.76}&	\textbf{9.97 } \\ 
Jigsaw (BERT)         & Medium  & 17.10&	16.44&	11.28&		11.80&	12.23&	\textbf{7.45}&	\textbf{7.66}  \\ 
(768, [144,277$,$ 15,294]) & High  & 7.19&	6.81&	4.84&		4.85&	3.43&	\textbf{0.78}&	\textbf{1.02} \\ %
\bottomrule
\end{tabular}}
\end{sc}
\vspace{-0.1in}
    \end{small}
\end{center}
\vskip -0.1in %
\end{table*}

\rev{
\begin{table}[!t]
\vspace{-0.1in}
\caption{The last/best epoch clean test accuracies (\%) when training with high-level noise defined in Table~\ref{tab:exp2}.}
\label{tab:exp_downstream}
\begin{center}
    \begin{small}
    \begin{sc}
\resizebox{0.8\linewidth}{!}{
\begin{tabular}{c |  c c c c}
\toprule
\multirow{2}{*}{Method} & \multicolumn{2}{c}{AG's news} & \multicolumn{2}{c}{DBpedia} \\
 & Last  & Best & Last  & Best \\
\midrule\midrule
HOC \cite{zhu2021clusterability} &  82.17 & 83.08 &  91.06 & 91.06\\
Ours-A-TV & \textbf{85.01} &  \textbf{85.17} &  \textbf{97.71} & \textbf{97.77}\\
\bottomrule
\end{tabular}}
\end{sc}
\vspace{-0.1in}
    \end{small}
\end{center}
\vskip -0.1in %
\end{table}
}

We evaluate our approaches on datasets with possibly lower-quality features in this section. The evaluation metric is the \textit{average total variation} between the true $\mT$ and the estimated $\hat \mT$ \cite{zhang2021learning} as Eqn.~(\ref{eq:T_error}).

\textbf{Baselines~}
We mainly compare our methods with three baselines: T-revision ({\sc T-Rev}) \cite{xia2019anchor}, Confident Learning ({\sc CL}) \cite{northcutt2021confident}, and the clusterability-based approach \cite{zhu2021clusterability} ({\sc HOC}). 
Some recent learning-based methods \cite{li2021provably,zhang2021learning} jointly optimize $\mT$ and the model during learning. We find their methods are unstable compared to the above baseline approaches on non-image datasets and defer their results to Appendix~\ref{appendix:moreexp}.

\textbf{Our approach~}
To evaluate each component of our approach, we test four variants: {\sc Ours-$\mX$-KL}, {\sc Ours-$\mX$-TV}, {\sc Ours-$\mA$-KL}, and {\sc Ours-$\mA$-TV}.
Prefix {\sc Ours-$\mX$} indicates that we directly use the original input features and substitute the soft cosine similarity for the original hard one employed by HOC. This setting checks the performance of ignoring correlations among different feature variables.
Prefix {\sc Ours-$\mA$} indicates that we firstly transform $\mX$ to $\mA$ as Section~\ref{sec:proxy_W} then apply soft cosine similarity.
Suffixes -KL and -TV denote using the $f$-mutual information when $f(v)=v\log(v)$ and $f(v)=\frac{1}{2}|v-1|$, respectively.
The matrix $\mW$ for soft cosine similarity is: $W_{\mu\mu} = \phi(I_f(X_\mu,\widetilde Y)),W_{\mu\nu} = 0, \forall \nu\ne\mu$, where $\phi(x)$ is an order-preserving activation function. In our evaluations, we set $\phi(x)=[x]_0^1$ for tabular benchmarks and $\phi(x)=[\log(x)]_0^1$ for natural language benchmarks, where $[x]_0^1$ represents normalizing $x$ to range $(0,1]$.

\textbf{Datasets and models~}
We examine our approaches on two different application domains other than images: the tabular benchmarks including $7$ tabular datasets from the UCI machine learning repository \cite{dua2017uci} and $4$ natural language benchmarks including AG's news \cite{NIPS2015_250cf8b5}, DBpedia \cite{dbpedia}, Yelp-5 \cite{yelp}, and Jigsaw \cite{Jigsaw}. We use the raw features for tabular benchmarks. \rev{Following the same preprocessing procedure as \citet{wang2022understanding}, we use a pre-trained BERT model \cite{devlin-etal-2019-bert} to extract 768 dimensional embedding vectors for natural language benchmarks.}

\textbf{Noise type~}
We synthesize the noisy data distribution by injecting class-dependent noise. Particularly, on tabular benchmarks, we test both the symmetric and the asymmetric label noise in Table~\ref{tab:exp1}. On natural language benchmarks, we randomly generate the diagonal-dominant label noise following the Dirichlet distribution. Particularly, suppose the average noise rate is $e$. For each row of $\mT$, We randomly sample a diagonal element following $T_{ii} = e+\textsf{Unif}(-0.05,0.05)$ and set the off-diagonal elements following $\textsf{Dir}(\bm 1)$, where $\textsf{Unif}(a,b)$ denotes the uniform distribution bounded by $(a,b)$, $\textsf{Dir}(\bm 1)$ denotes the Dirichlet distribution with parameter $\bm 1:=[1,\cdots,1]$ ($K-1$ values).

\subsection{Evaluation on Tabular Benchmarks}
Table~\ref{tab:exp1} shows the performance comparisons on tabular datasets. By counting the number of \textbf{bold} (top-2 performance) results, we know all the four variants of our proposed method perform better than three baselines statistically. Additionally, based on the counting results, {\sc Ours-$\bm X$} wins in $16$ settings while {\sc Ours-$\bm A$} wins in $20$ settings, indicating decoupling different parts by eigen decomposition is not statistically significantly useful. This may be due to the property of tabular data: the original correlation in $\mX$ may not be strong and the decoupling operations will involve extra errors and make the results even worse.

\vspace{-2ex}
\subsection{Evaluation on Natural Language Benchmarks}
We also evaluate our methods on more sophisticated text classifications tasks.
We use a heuristic method to set the average noise rate $e$. The aim is to make the ratio between diagonal elements and off-diagonal elements of $\mT$ consistent. According to our rule in the caption of Table~\ref{tab:exp2}, the low, medium, and high noise rates for binary, $5$-class, and $10$-class classifications are $(0.11,0.2,0.4)$, $(0.2,0.33,0.57)$, and $(0.27,0.43,0.67)$, respectively, which align with the general cognition of the community \cite{cheng2021learningsieve}.
Comparing Table~\ref{tab:exp2} with Table~\ref{tab:exp1}, we find the direct reweighting by $f$-mutual information ({\sc Ours-$\mX$}) performs similarly to the hard cosine similarity adopted by HOC, while the information-theoretic reweighting after projecting $\mX$ to its eigen space ({\sc Ours-$\mA$}) performs consistently better than other methods. Intuitively, the correlations of BERT embeddings are more stronger than those of tabular data. Thus we observe a clear performance improvement by removing correlations.
Besides, it is interesting to see \textit{the estimation error may decrease for higher noise rate setting}.We conjecture the reasons may comprise: 1) The original dataset may be noisy, e.g., the original Yelp dataset contains lots of noisy reviews \cite{luca2016reviews}. Consider a binary classification with inherent 20\% noise. Then adding 10\% (low) noise will make the average noise rate to $0.26$, where the gap between the real noise rate and the hypothesized noise rate is $0.26-0.1=0.16$. Similarly, the gap of adding 40\% (high) noise is $0.44-0.4=0.04$. Therefore, even though $\mT$ is accurately estimated, the absolute error under our current metric will be higher for the {\sc High}-noise case.
2) As analyzed in Section~\ref{sec:failures}, the error of random guess for low-noise (10\%) and high-noise (40\%) settings are $0.4$ and $0.1$, respectively, indicating a small error may cause more severe problems in higher-noise settings.
We leave more detailed discussions to Appendix~\ref{appendix:discuss}.

\rev{\textbf{Downstream learning error~} \citet{liu2021can} showed the additional learning risk is positively related to estimation error. To further consolidate the estimation error reduction of our method, we feed the estimated $\mT$ into forward loss correction \cite{patrini2017making} and check the clean test accuracy.
Table~\ref{tab:exp_downstream} shows our approach significantly improves both the best and last epoch test accuracy by simply changing the $\mT$ in loss correction from HOC estimates to ours.

}

\vspace{-5pt}
\section{Conclusions}
\vspace{-5pt}
This work has studied the problem of estimating noise transition matrix on application domains apart from images. We have proposed an information-theoretic approach to down-weight the less informative parts of features with only noisy labels for tasks with lower-quality features. Future directions include implementing and delivering this approach in real-world label noise settings, e.g., long-tail and open-set \cite{wei2022open,hu2019weakly}.

\textbf{Acknowledgment}
This work is partially supported by the National Science Foundation (NSF) under grants IIS-2007951, IIS-2143895 and CCF-2023495.

\clearpage
\newpage
\bibliography{ref}
\bibliographystyle{icml2022}

\newpage
\appendix
\onecolumn

\section*{Appendix}

We show more theoretical details in Section~\ref{appendix:thm} and more emperical details in Section~\ref{appendix:exp}. Particuarly,
\squishlist
\item Section~\ref{appendix:f-div} numerates some popular $f$-divergence functions and the corresponding optimal variational-conjugate pair $(g^*,f^*)$.
\item Section~\ref{appendix:tv} shows the order-preserving property of using total variation.
\item Section~\ref{appendix:kl} shows the order-preserving property of using KL divergence.
\item Section~\ref{appendix:rationale} explains why we build our method on HOC.
\item Section~\ref{appendix:moreexp} compares two more baselines.
\item Section~\ref{appendix:discuss} discuss an interesting observation shown in Table~\ref{tab:exp2} that high-noise settings may have lower errors.
\squishend

\section{Theorems}\label{appendix:thm}

\subsection{Common $f$-Divergence Functions}\label{appendix:f-div}
Following \cite{nowozin2016f,wei2020optimizing}, we show some common $f$-divergence functions in Table~\ref{table:f_div-full},
\begin{table}[!htb]
\caption{List of popular $f$-divergences together with generator functions $f(v)$, optimal variational functions $g^*$ and optimal conjugate functions $f^*$.}
\small
\begin{center}
\begin{tabular}{ l l l l l} 
 \toprule
 Name  & $f(v)$ & $g^*$ & $\text{dom}_{f^*}$ & $f^*(u)$\\
 \midrule
 Total Variation   & $\dfrac{1}{2} |v-1|$ & $\dfrac{1}{2}\sign\left(\dfrac{p(v)}{q(v)}-1\right)$ & $u\in [-\dfrac{1}{2}, \dfrac{1}{2}]$ & $u$ \\
 KL   & $v\log v$ &$1+\log{\dfrac{p(v)}{q(v)}}$ &$\mathbb{R}$ & $e^{u-1}$\\
 Jenson-Shannon  & $-(1+v)\log\frac{1+v}{2} + v\log v$ &$\log{\dfrac{2p(v)}{p(v)+q(v)}}$ & $u<\log{2}$  & $-\log{(2-e^{u})}$\\
 Squared Hellinger  &$(\sqrt{v}-1)^2$ & $1-\sqrt{\dfrac{q(v)}{p(v)}}$ & $u<1$ & $\dfrac{u}{1-u}$\\
 Pearson $\mathcal X^2$   & $(1-v)^2$ & $2\left(\dfrac{p(v)}{q(v)}-1\right)$ & $\mathbb{R}$ & $\dfrac{1}{4}u^2+u$ \\
 Neyman  $\mathcal X^2$  &$\frac{(1-v)^2}{v}$ &$1-\left(\dfrac{q(v)}{p(v)}\right)^2$ & $u<1$ & $2-2\sqrt{1-u}$\\
 Reverse KL  & $-\log v$ & $-\dfrac{q(v)}{p(v)}$ & $\mathbb{R}_-$  &$-1-\log{(-u)}$ \\
 \bottomrule
\end{tabular}
\end{center}
\label{table:f_div-full}
\end{table}

\subsection{Total-Variation}\label{appendix:tv}

\subsubsection{Proof for Lemma~\ref{lem:tv_gap}}
\begin{proof}

Consider TV, we have:
{\small\begin{align*}
    {\textsf{VD}}_f(\tilde g^*) 
    = & \E_{V\sim P}[\tilde g^*(V)] - \E_{V\sim Q}[f^*(\tilde g^*(V))] \\
    = & \frac{1}{2} \left[\E_{V\sim P}\left[ \sign\left( \frac{\tilde p(V)}{\tilde q(V)} -1 \right) \right] - \E_{V\sim Q}\left[ \sign\left( \frac{\tilde p(V)}{\tilde q(V)} -1 \right) \right] \right]\\
    = & \frac{1}{2} \int_z \sum_{i\in\{1,2\}} \left[\PP(Z=z, Y=i)-\PP(Z=z) \PP(Y=i)\right] \cdot  \sign \left[\PP(Z=z, \widetilde Y=i)-\PP(Z=z) \PP(\widetilde  Y=i)\right] ~dz\\
    = & \frac{1}{2}  \left[\PP(Z=z, Y=1)-\PP(Z=z) \PP(Y=1)\right] \cdot  \sign \left[\PP(Z=z, \widetilde Y=1)-\PP(Z=z) \PP(\widetilde  Y=1)\right] \\
    & + \frac{1}{2}\left[\PP(Z=z, Y=2)-\PP(Z=z) \PP(Y=2)\right] \cdot  \sign \left[\PP(Z=z, \widetilde Y=2)-\PP(Z=z) \PP(\widetilde  Y=2)\right] ~dz\\
    = & \frac{1}{2}  \left[\PP(Z=z, Y=1)-\PP(Z=z) \PP(Y=1)\right] \\
     &\cdot \sign \left[\sum_{j\in\{1,2\}} \left(\PP( \widetilde Y=1 | Z=z, Y=j)\PP( Z=z, Y=j)-\PP(Z=z) \PP(\widetilde  Y=1 | Y=j)\PP(Y=j) \right)\right] \\
    & + \frac{1}{2}\left[\PP(Z=z, Y=2)-\PP(Z=z) \PP(Y=2)\right] \\
    & \cdot  \sign \left[\sum_{j\in\{1,2\}} \left(\PP( \widetilde Y=2 | Z=z, Y=j)\PP(Z=z,Y=j)-\PP(Z=z) \PP(\widetilde  Y=2|Y=j) \PP(Y=j)\right)\right] ~dz.
\end{align*}}
Note (assume class-dependent label noise)

{\small\begin{align*}
    & \sign \left[\sum_{j\in\{1,2\}} \left(\PP( \widetilde Y=1 | Z=z, Y=j)\PP( Z=z, Y=j)-\PP(Z=z) \PP(\widetilde  Y=1 | Y=j)\PP(Y=j) \right)\right] \\
  = & \sign \left[ (1-e_1)\PP( Z=z, Y=1) + e_2 \PP( Z=z, Y=2) -  (1-e_1)\PP(Z=z)\PP(Y=1) - e_2 \PP(Z=z)\PP(Y=2)\right] \\
  = & \sign \left[ (1-e_1) \left(\PP( Z=z, Y=1) - \PP(Z=z)\PP(Y=1) \right) + e_2 \left( \PP( Z=z, Y=2) - \PP(Z=z)\PP(Y=2)\right)\right] \\
  = & \sign \left[ (1-e_1) \left(\PP( Z=z, Y=1) - \PP(Z=z)\PP(Y=1) \right) + e_2 \left( \PP(Z=z) - \PP( Z=z, Y=1) -  \PP(Z=z)(1-\PP(Y=1))\right)\right] \\
  = & \sign  (1-e_1 - e_2)  \cdot \sign(\PP( Z=z, Y=1) - \PP(Z=z)\PP(Y=1)).
\end{align*}}
Thus
\begin{align*}
    & {\textsf{VD}}_f(\tilde g^*) \\
  = &  \frac{1}{2} \int_z  \left[\PP(Z=z, Y=1)-\PP(Z=z) \PP(Y=1)\right]\cdot  \sign  (1-e_1 - e_2)  \cdot \sign(\PP( Z=z, Y=1) - \PP(Z=z)\PP(Y=1)) \\
    & + \left[\PP(Z=z, Y=2)-\PP(Z=z) \PP(Y=2)\right] \cdot  \sign  (1-e_1 - e_2)  \cdot \sign(\PP( Z=z, Y=2) - \PP(Z=z)\PP(Y=2)) ~dz.
\end{align*}
When $\sign(1-e_1-e_2) = 1$, i.e. $e_1 + e_2 < 1$, we have
\begin{align*}
    & {\textsf{VD}}_f(\tilde g^*) \\
  = &  \frac{1}{2} \int_z \sum_{i\in\{1,2\}} \left[\PP(Z=z, Y=i)-\PP(Z=z) \PP(Y=i)\right]\cdot   \sign(\PP( Z=z, Y=i) - \PP(Z=z)\PP(Y=i)) ~dz \\
  = & {\textsf{VD}}_f(g^*).
\end{align*}
\end{proof}

\subsubsection{Proof for Theorem~\ref{thm:tvrobust}}
\begin{proof}
Recall that, to show an $f$-mutual information metric is $\epsilon$-order-preserving under label noise, we need to study how $\widetilde{\textsf{VD}}_f(\tilde g^*)$ differs from the order of  ${\textsf{VD}}_f(g^*)$. 

For total variation, with Lemma~\ref{lem:linear} and Lemma~\ref{lem:tv_gap}, we know
\[
\widetilde{\textsf{VD}}_{\text{TV}}(\tilde g^*) = (1-e_1 - e_2) {\textsf{VD}}_{\text{TV}}(\tilde g^*) = (1-e_1 - e_2) {\textsf{VD}}_{\text{TV}}(g^*).
\]
Therefore, when $e_1+e_2 < 1$, $\widetilde{\textsf{VD}}_{\text{TV}}(\tilde g^*)$ always preserves the order of ${\textsf{VD}}_{\text{TV}}(g^*)$, indicating the total-variation-based mutual information is $0$-order-preserving under class-dependent label noise.
\end{proof}

\subsection{KL Divergence}\label{appendix:kl}

The definition of MI is
{\small\begin{align*}
  &  I(Z,\widetilde Y) \\
= & \sum_{j\in \{1,2\}} \int_{z} \PP(Z=z,\widetilde Y=j)  \log \left(\frac{\PP(Z=z,\widetilde Y=j)}{\PP(Z=z)\PP(\widetilde Y=j)}\right) dz \\
= & \sum_{j\in \{1,2\}} \int_{z} \PP(Z=z,\widetilde Y=j)  \log \left(\PP(Z=z,\widetilde Y=j)\right) dz  - \sum_{j\in \{1,2\}} \int_{z} \PP(Z=z,\widetilde Y=j) \left( \log \left(\PP(Z=z\right) + \log \left(\PP(\widetilde Y=j)\right)\right) dz  \\
= & \underbrace{\sum_{j\in \{1,2\}} \int_{z} \PP(Z=z,\widetilde Y=j)  \log \left(\PP(Z=z,\widetilde Y=j)\right) dz}_{\text{\textbf{Term-1:}~}-H(Z,\widetilde Y)}  -  \underbrace{\int_{z} \PP(Z=z) \log \PP(Z=z) ~dz}_{\text{\textbf{Term-2:}~}-H(Z)} - \underbrace{\sum_{j\in \{1,2\}}\PP(\widetilde Y=j) \log \PP(\widetilde Y=j) dz  }_{\text{\textbf{Term-3:}~}-H(\widetilde Y)},
\end{align*}}
where
\begin{align*}
    -H(Z,\widetilde Y)= &\sum_{j\in \{1,2\}} \int_{z} \PP(Z=z,\widetilde Y=j)  \log \PP(Z=z,\widetilde Y=j) dz \\
   = & \int_{z} \PP(Z=z,\widetilde Y=1)  \log \PP(Z=z,\widetilde Y=1) + \PP(Z=z,\widetilde Y=2)  \log \PP(Z=z,\widetilde Y=2)~ dz.
\end{align*}

We first decouple term-1. Note
{\begin{align*}
     &   \int_{z}  \PP(Z=z,\widetilde Y=1)  \log \PP(Z=z,\widetilde Y=1) dz \\
     =&  \int_{z} \left[ \PP(\widetilde Y=1|Z=z,Y=1)\PP(Z=z,Y=1) +  \PP(\widetilde Y=1|Z=z,Y=2)\PP(Z=z,Y=2) \right] \\
     & \cdot \log \left[ \PP(\widetilde Y=1|Z=z,Y=1)\PP(Z=z,Y=1) +  \PP(\widetilde Y=1|Z=z,Y=2)\PP(Z=z,Y=2) \right] dz \\
    =&  \int_{z} \left[ (1-e_1)\PP(Z=z,Y=1) + e_2\PP(Z=z,Y=2) \right] \\
     & \cdot \log \left[(1-e_1)\PP(Z=z,Y=1) + e_2\PP(Z=z,Y=2)  \right] dz \\
    =&  \int_{z} \left[ (1-e_1-e_2)\PP(Z=z,Y=1) + e_2\PP(Z=z) \right] \\
     & \cdot \log \left[(1-e_1-e_2)\PP(Z=z,Y=1) + e_2\PP(Z=z)  \right] dz \\
    =&  \int_{z} \left[ (1-e_1-e_2)\PP(Z=z,Y=1) + e_2\PP(Z=z) \right] \log \PP(Z=z,Y=1)  \\
     & + \left[ (1-e_1-e_2)\PP(Z=z,Y=1) + e_2\PP(Z=z) \right]  \log \left[ \frac{(1-e_1-e_2)\PP(Z=z,Y=1) + e_2\PP(Z=z)}{\PP(Z=z,Y=1)}  \right] dz \\
    =&  \int_{z} \left[ (1-e_1-e_2)\PP(Z=z,Y=1) + e_2\PP(Z=z) \right] \log \PP(Z=z,Y=1)  \\
     & + \left[ (1-e_1-e_2)\PP(Z=z,Y=1) + e_2\PP(Z=z) \right]  \log \left[1-e_1 + e_2 \frac{\PP(Z=z,Y=2)}{\PP(Z=z,Y=1)} \right] dz.
\end{align*}}

Let $\alpha = \PP(Z=z,Y=1) / \PP(Z=z,Y=2) \in [0,+\infty)$ (note $\alpha$ is actually a function of $(Z,Y)$). 
Then
\begin{align*}
    &   \int_{z}  \PP(Z=z,\widetilde Y=1)  \log \PP(Z=z,\widetilde Y=1) dz \\
    =&  \int_{z} (1-e_1-e_2)\PP(Z=z,Y=1)  \log \PP(Z=z,Y=1) ~dz   + \int_{z} e_2\PP(Z=z) \left[\log \alpha + \log \PP(Z=z,Y=2) \right]   \\
     & + \left[ (1-e_1-e_2)\alpha\PP(Z=z,Y=2) + e_2\PP(Z=z) \right]  \log \left(1-e_1 +  \frac{e_2}{\alpha} \right) dz 
\end{align*}
and
\begin{align*}
    &   \int_{z}  \PP(Z=z,\widetilde Y=2)  \log \PP(Z=z,\widetilde Y=2) dz \\
    =&  \int_{z} (1-e_1-e_2)\PP(Z=z,Y=2)  \log \PP(Z=z,Y=2) ~dz  + \int_{z} e_1\PP(Z=z) \log \PP(Z=z,Y=2)  \\
     & + \left[ (1-e_1-e_2)\PP(Z=z,Y=2) + e_1\PP(Z=z) \right]  \log \left(1-e_2 +  {e_1}{\alpha} \right) dz 
\end{align*}

Thus
{\begin{align*}
    &\sum_{j\in \{1,2\}} \int_{z} \PP(Z=z,\widetilde Y=j)  \log \PP(Z=z,\widetilde Y=j) dz \\
   = &(1-e_1-e_2)  \sum_{i\in\{1,2\}}\int_{z} \PP(Z=z,Y=i)  \log \PP(Z=z,Y=i) ~dz  \\
   & + \int_{z} e_2\PP(Z=z)\log \alpha + (e_1 + e_2) \PP(Z=z) \log \PP(Z=z,Y=2)    \\
   & + (1-e_1-e_2)\PP(Z=z,Y=2) \left[\alpha \log \left(1-e_1 +  \frac{e_2}{\alpha} \right) + \log(1-e_2+e_1\alpha)\right] \\
   & + \PP(Z=z) \left[e_1\log(1-e_2+e_1\alpha) + e_2 \log \left(1-e_1 +  \frac{e_2}{\alpha} \right)\right] ~dz\\ 
   = &(1-e_1-e_2)  \sum_{i\in\{1,2\}}\int_{z} \PP(Z=z,Y=i)  \log \PP(Z=z,Y=i) ~dz  \\
   & + \int_{z} e_2\PP(Z=z)\log \alpha - (e_1+e_2)\PP(Z=z)\log(\alpha+1) + (e_1 + e_2) \PP(Z=z) \log \PP(Z=z)    \\
   & + \frac{(1-e_1-e_2)}{\alpha+1}\PP(Z=z) \left[\alpha \log \left(1-e_1 +  \frac{e_2}{\alpha} \right) + \log(1-e_2+e_1\alpha)\right] \\
   & + \PP(Z=z) \left[e_1\log(1-e_2+e_1\alpha) + e_2 \log \left(1-e_1 +  \frac{e_2}{\alpha} \right)\right] ~dz\\ 
   & = (1-e_1-e_2)  \sum_{i\in\{1,2\}}\int_{z} \PP(Z=z,Y=i)  \log \PP(Z=z,Y=i) ~dz  \qquad \qquad \quad \text{\textbf{(Term 1.1)}}\\
   & + \int_z (e_1 + e_2) \PP(Z=z) \log \PP(Z=z) + \PP(Z=z)\Delta_{\textsf{Bias}}(\alpha, e_1, e_2) ~ dz,  \qquad \qquad \text{\textbf{(Term 1.2)}}
\end{align*}}
where in Term 1.2:
{\small\begin{equation}\label{eqn:delta_bias}
\begin{split}
        \Delta_{\textsf{Bias}}(\alpha, e_1, e_2) = &e_2\log \alpha - (e_1+e_2) \log(\alpha+1)      + \frac{(1-e_1-e_2)}{\alpha+1}  \left[\alpha \log \left(1-e_1 +  \frac{e_2}{\alpha} \right) + \log(1-e_2+e_1\alpha)\right] \\
   & +  \left[e_1\log(1-e_2+e_1\alpha) + e_2 \log \left(1-e_1 +  \frac{e_2}{\alpha} \right)\right].
\end{split}
\end{equation}}

In Term 1.1, recalling $\alpha = \PP(Z=z,Y=1) / \PP(Z=z,Y=2)$, we have
{\small\begin{align*}
    & (1-e_1-e_2)  \sum_{i\in\{1,2\}}\int_{z} \PP(Z=z,Y=i)  \log \PP(Z=z,Y=i) ~dz \\
 =  & (1-e_1-e_2)  \int_{z} \frac{\PP(Z=z)}{\alpha + 1}  \log \frac{\PP(Z=z)}{\alpha + 1} + \frac{\PP(Z=z)\alpha}{\alpha + 1} \log \frac{\PP(Z=z)\alpha}{\alpha + 1} ~dz \\
 =  & (1-e_1-e_2)  \int_{z} \frac{\PP(Z=z)}{\alpha + 1}  \log {\PP(Z=z)} + \frac{\PP(Z=z)}{\alpha + 1}\log \frac{1}{\alpha + 1}  + \frac{\PP(Z=z)\alpha}{\alpha + 1} \log {\PP(Z=z)} + \frac{\PP(Z=z)\alpha}{\alpha + 1}\log\frac{\alpha}{\alpha + 1} ~dz \\
 =  & (1-e_1-e_2)  \int_{z}{\PP(Z=z)}  \log {\PP(Z=z)} ~dz  + (1-e_1-e_2) \int_{z}{\PP(Z=z)} \left[ \frac{\alpha}{\alpha+1}\log\frac{\alpha}{\alpha+1} + \frac{1}{\alpha+1}\log\frac{1}{\alpha+1} \right]~dz.
\end{align*}}

Denote the effective part of MI by
\begin{equation}\label{eqn:delta_mi}
    \Delta_{\textsf{MI}}(\alpha, e_1, e_2) = (1-e_1-e_2)  \left[ \frac{\alpha}{\alpha+1}\log\frac{\alpha}{\alpha+1} + \frac{1}{\alpha+1}\log\frac{1}{\alpha+1} \right]. 
\end{equation}

We have 
\begin{align*}
     -H(Z,\widetilde Y) = &\sum_{j\in \{1,2\}} \int_{z} \PP(Z=z,\widetilde Y=j)  \log \PP(Z=z,\widetilde Y=j) dz \\
   & =   \int_z \PP(Z=z) \log \PP(Z=z) ~dz  + \int_z \PP(Z=z) \left[\Delta_{\textsf{MI}}(\alpha, e_1, e_2) + \Delta_{\textsf{Bias}}(\alpha, e_1, e_2) \right] ~ dz,
\end{align*}
and
\begin{align*}
  &  I(Z,\widetilde Y) =  \int_z \PP(Z=z) \left[\Delta_{\textsf{MI}}(\alpha, e_1, e_2) + \Delta_{\textsf{Bias}}(\alpha, e_1, e_2) \right] ~ dz + {H(\widetilde Y)}.
\end{align*}

Define
$
\Delta_{\textsf{Bias,MI}}(\alpha, e_1, e_2) = \Delta_{\textsf{MI}}(\alpha, e_1, e_2) +  \Delta_{\textsf{Bias}}(\alpha, e_1, e_2).
$
Then 
{\small\begin{align*}
&\Delta_{\textsf{Bias,MI}} (\alpha, e_1, e_2) \\
= & \Delta_{\textsf{MI}} (\alpha, e_1, e_2)+  \Delta_{\textsf{Bias}}(\alpha, e_1, e_2) \\
= &   (1-e_1-e_2)  \left[ \frac{\alpha}{\alpha+1}\log\frac{\alpha}{\alpha+1} + \frac{1}{\alpha+1}\log\frac{1}{\alpha+1} \right] +   e_2\log \alpha - (e_1+e_2)\log(\alpha+1)  \\
   & + \frac{(1-e_1-e_2)}{\alpha+1} \left[\alpha \log \left(1-e_1 +  \frac{e_2}{\alpha} \right) + \log(1-e_2+e_1\alpha)\right] \\
   & +  \left[e_1\log(1-e_2+e_1\alpha) + e_2 \log \left(1-e_1 +  \frac{e_2}{\alpha} \right)\right]\\
= &   (1-e_1-e_2)  \left[ \frac{\alpha}{\alpha+1}\log \alpha + \log\frac{1}{\alpha+1} \right] +   e_2\log \alpha + (e_1+e_2)\log\frac{1}{\alpha+1} + \\
   & \left[ \frac{(1-e_1-e_2)\alpha}{\alpha+1} + e_2\right] \log \left(1-e_1 +  \frac{e_2}{\alpha} \right)  + \left[\frac{(1-e_1-e_2)}{\alpha+1} +e_1 \right] \log(1-e_2+e_1\alpha) \\
= &  \left[ \frac{(1-e_1-e_2)\alpha}{\alpha+1} + e_2 \right]\log \alpha + \log \frac{1}{\alpha+1} +\\
   & \left[ \frac{(1-e_1-e_2)\alpha}{\alpha+1} + e_2\right] \log \left(1-e_1 +  \frac{e_2}{\alpha} \right)  + \left[\frac{(1-e_1-e_2)}{\alpha+1} +e_1 \right] \log(1-e_2+e_1\alpha) \\
= &  \log \frac{1}{\alpha+1}
+\frac{1}{\alpha+1}\left[ {(1-e_1-e_2)\alpha} + e_2(\alpha+1)\right] \log \left(\alpha(1-e_1) + e_2  \right) \\
&+ \frac{1}{\alpha+1} \left[(1-e_1-e_2) +e_1(\alpha+1) \right] \log(1-e_2+e_1\alpha) \\
= & \left[ \frac{\alpha(1-e_1) + e_2}{\alpha+1} 
+
\frac{(\alpha+1) - (\alpha(1-e_1) + e_2)}{\alpha+1} 
\right]
\log \frac{1}{\alpha+1} +\frac{1}{\alpha+1}\left[\alpha(1-e_1) + e_2 \right] \log \left(\alpha(1-e_1) + e_2  \right)\\
   &   + \frac{1}{\alpha+1} \left[(\alpha+1) - (\alpha(1-e_1) + e_2)  \right] \log((\alpha+1) - (\alpha(1-e_1) + e_2) ) \\
= &\frac{\alpha(1-e_1) + e_2}{\alpha+1} \log \frac{\alpha(1-e_1) + e_2}{\alpha+1}+ \left[1- \frac{ \alpha(1-e_1) + e_2}{\alpha+1} \right] \log \left[1- \frac{ \alpha(1-e_1) + e_2}{\alpha+1} \right].
\end{align*}}
Note 
\[
\frac{\alpha(1-e_1) + e_2}{\alpha+1} = (1-e_1-e_2) \cdot  \frac{\alpha}{\alpha+1}  + e_2.
\]
Let $\beta = \alpha/(1+\alpha) \in [0,1)$. Note $\beta$ is a function of $z$. We drop notation $z$ for ease of notation.
Then 
\begin{equation*}
\begin{split}
        \Delta_{\textsf{Bias,MI}} (\beta, e_1, e_2) = & \left((1-e_1-e_2) \beta  + e_2 \right) \log \left((1-e_1-e_2) \beta   + e_2 \right) \\
    & + \left[1- \left((1-e_1-e_2) \beta  + e_2 \right) \right] \log \left[1- \left((1-e_1-e_2) \beta  + e_2 \right) \right].
\end{split}
\end{equation*}

The bias caused by label noise is
\begin{equation}\label{eqn:delta_bias_beta}
\begin{split}
    \Delta_{\textsf{Bias}}(\beta, e_1, e_2) 
    = & \Delta_{\textsf{Bias,MI}} (\beta, e_1, e_2) -  \Delta_{\textsf{MI}} (\beta, e_1, e_2) \\  
    = & \left((1-e_1-e_2) \beta  + e_2 \right) \log \left((1-e_1-e_2) \beta   + e_2 \right) \\
    & + \left[1- \left((1-e_1-e_2) \beta  + e_2 \right) \right] \log \left[1- \left((1-e_1-e_2) \beta  + e_2 \right) \right] \\
    & - (1-e_1-e_2)  \left[ \beta\log\beta + (1-\beta)\log(1-\beta)\right].
\end{split}
\end{equation}

To get $\argmax_{\beta \in [0,1)}$, we check the first derivative:
\begin{align*}
    \frac{\partial \Delta_{\textsf{Bias}}(\beta, e_1, e_2)}{\partial \beta} = (1-e_1-e_2) \left[ \log\frac{(1-e_1-e_2)\beta + e_2}{1- (1-e_1-e_2)\beta - e_2} - \log\frac{\beta}{1-\beta} \right]. 
\end{align*}
Let $\pdv*{ \Delta_{\textsf{Bias}}(\beta, e_1, e_2)}{\beta} = 0$, we have
\[
\beta^* = \frac{e_2}{e_1+e_2}.
\]
By checking $\pdv*[2]{\Delta_{\textsf{Bias}}(\beta, e_1, e_2)}{\beta}$, we can find that $\Delta_{\textsf{Bias}}(\beta, e_1, e_2) $ is increasing when $\beta \in [0,\beta^*]$, and decreasing when $\beta \in [\beta^*,1]$. Thus $\beta^* = \nicefrac{e_2}{(e_1+e_2)}$ is the global maximum and the upper bound for $\Delta_{\textsf{Bias}}(\beta, e_1, e_2) $ is
\[
\Delta_{\textsf{Bias}}(\beta, e_1, e_2)  \le e_1\log e_1 + e_2 \log e_2 - (e_1+e_2)\log(e_1 + e_2).
\]
Assume $\beta \in [\underline{\beta}, \bar \beta]$ in practice. The lower bound for  $\Delta_{\textsf{Bias}}(\beta, e_1, e_2) $ is
\[
\Delta_{\textsf{Bias}}(\beta, e_1, e_2)  \ge \min(\Delta_{\textsf{Bias}}(\underline{\beta}, e_1, e_2), \Delta_{\textsf{Bias}}(\bar\beta, e_1, e_2)).
\]
A looser bound that holds for all the possible $\beta \in [0,1)$ is:
\[
\Delta_{\textsf{Bias}}(\beta, e_1, e_2)  \ge \min(e_1 \log e_1 + (1-e_1)\log (1-e_1), e_2 \log e_2 + (1-e_2)\log (1-e_2)).
\]

Note (when $e_1=e_2 = 0$)
\begin{align*}
  &  I(Z,Y) =  \int_z \PP(Z=z) \left[\Delta_{\textsf{MI}}(\alpha,0,0) \right] ~ dz + {H( Y)},
\end{align*}
and $\Delta_{\textsf{MI}}(\beta,e_1,e_2) = (1-e_1-e_2)\Delta_{\textsf{MI}}(\alpha,0,0)$.

Hence (note $\beta$ is actually a function of $Z$),
\begin{align*}
   I(Z,\widetilde Y)  = & \int_z \PP(Z=z) \left[\Delta_{\textsf{MI}}(\alpha, e_1, e_2) + \Delta_{\textsf{Bias}}(\alpha, e_1, e_2) \right] ~ dz + {H(\widetilde Y)} \\
   = & \int_z \PP(Z=z) \left[ (1-e_1-e_2) \Delta_{\textsf{MI}}(\beta, 0, 0) + \Delta_{\textsf{Bias}}(\beta, e_1, e_2) \right] ~ dz + {H(\widetilde Y)} \\
   = &(1-e_1-e_2) \int_z \PP(Z=z)   \Delta_{\textsf{MI}}(\beta, 0, 0) ~ dz + \int_z \Delta_{\textsf{Bias}}(\beta, e_1, e_2)  ~dz  + {H(\widetilde Y)} \\
   = &(1-e_1-e_2) I(Z,Y)  + \int_z \Delta_{\textsf{Bias}}(\beta, e_1, e_2)  ~dz  - (1-e_1-e_2)H(Y) + {H(\widetilde Y)} \\
   = &(1-e_1-e_2) I(Z,Y)  + C(e_1,e_2,Y,\widetilde Y) + \Delta_Z(e_1,e_2),
\end{align*}
where 
\[
 C(e_1,e_2,Y,\widetilde Y) = \min(e_1 \log e_1 + (1-e_1)\log (1-e_1), e_2 \log e_2 + (1-e_2)\log (1-e_2)) - (1-e_1-e_2)H(Y) + {H(\widetilde Y)}
\]
is a constant for given $Y$ and $\widetilde Y$.
The other part is in the range
$\Delta_Z(e_1,e_2) \in [0,\textsf{Gap}_Z(e_1,e_2)]$,
and 
\begin{align*}
    \textsf{Gap}_Z(e_1,e_2) =&  e_1\log e_1 + e_2 \log e_2 - (e_1+e_2)\log(e_1 + e_2) \\
    & -\min(e_1 \log e_1 + (1-e_1)\log (1-e_1), e_2 \log e_2 + (1-e_2)\log (1-e_2)).
\end{align*}
Note $\Delta_Z(e_1,e_2)$ may be different for $Z_\mu$ and $Z_\nu$, $\mu\ne \nu$.

Therefore, when
\[
|I_f(Z_\mu;\widetilde Y) - I_f(Z_\nu;\widetilde Y)| > \textsf{Gap}_Z(e_1,e_2),
\]
we have
\[
\sign \left[I_f(Z_\mu;\widetilde Y) - I_f(Z_\nu;\widetilde Y) \right] = \sign \left[I_f(Z_\mu; Y) - I_f(Z_\nu; Y) \right], \forall \mu \in [d]\nu \in [d].
\]

Now we take a further look at the gap $\textsf{Gap}_Z(e_1,e_2) $.
Assume $e_1 \ge e_2 \Rightarrow e_2 = \delta e_1$, where $\delta \in [0,1]$. Then $H(e_1) \le H(e_2)$, and 
\begin{align*}
    \textsf{Gap}_Z(e_1,e_2) = &  e_1\log e_1 + e_2 \log e_2 - (e_1+e_2)\log(e_1 + e_2)-\min(H(e_1), H(e_2)) \\
    = & e_2 \log e_2 - (e_1 + e_2) \log(e_1 + e_2) - (1-e_1) \log (1-e_1) \\
    = & e_2 \log \frac{e_2}{e_1 + e_2} + e_1 \log \frac{1-e_1}{e_1+e_2} - \log (1-e_1) \\
    = & \delta e_1 \log\frac{\delta}{1+\delta} + e_1 \log \frac{1-e_1}{e_1(1+\delta)} - \log (1-e_1) \\
    = &  e_1 \left[ \delta\log\frac{\delta}{1+\delta} + \log \frac{1}{(1+\delta)}
    \right]  + e_1 \log \frac{(1-e_1)}{e_1} - \log (1-e_1) \\
    = &  e_1 \left[ \delta\log \delta - (1+\delta)\log (1+\delta)
    \right]  - (1-e_1) \log (1-e_1) - e_1\log e_1 \\
    = &  e_1 \left[ \delta\log \delta - (1+\delta)\log (1+\delta)
    \right]  + H(e_1) \\
\end{align*}

\textbf{Note:} We can also roughly estimate $ \delta\log \delta - (1+\delta)\log (1+\delta)$ by the best quadratic fit (rooted mean squared error $\approx 0.02$) and get
\[
\delta\log \delta - (1+\delta)\log (1+\delta) \approx 0.9124 \delta^2 - 2.14 \delta - 0.1202.
\]

\section{More Discussions}\label{appendix:exp}

\subsection{Rationale for building on HOC}\label{appendix:rationale}
The rationale for building our analyses on HOC is described as follows. Our major concern is that the learning-based approaches usually require more effort in tuning their hyperparameters. The effect of reweighing feature variables will be entangled with the training procedure, making things more complicated. On the other hand, the training-free approach seems to be more lightweight to employ the reweighing treatment. In particular, HOC consistently achieves lower estimation error, as shown in Figure~\ref{fig:failures}.

\subsection{More Experimental Results}\label{appendix:moreexp}

We only use one linear layer as the model for the learning-based methods since it can achieve satisfying performance when training with clean data.

We compare our methods with two recent works \cite{li2021provably,zhang2021learning} in Table~\ref{tab:appendix_more_exp}. Our method achieves an overall better performance than theirs. We search the learning rate from $[0.1, 0.01, 0.001, 0.0001, 0.00001]$ and report the best result for their methods. During experiments, we find their methods tend to be sensitive to hyperparameter settings.
\rev{On AG's news and Jigsaw, we find T-Vol and T-TV estimate $\mT \approx \mI$ for all three noise settings. Thus the low estimation error for the low-noise setting is more like a coincidence due to the trivial solution $\mI$.}

\begin{table*}[h]
\caption{The estimation error ($\times 100$) on natural language benchmarks. Feature dimension: $d$. Number of clean instances in class-$k$: $N_{k}$. {\sc [30k $\times$ 4]}: $N_1=N_2=N_3=N_4=$30k. Average noise rate follows $e={1}/{(1+r/\sqrt{K-1})}$. {\sc Low: $r=8$. Medium: $r=4$. High: $r=1.5$.} Top-2 of each row are \textbf{bold}.}
\label{tab:appendix_more_exp}
\vskip 0.1in %
\begin{center}
    \begin{small}
    \begin{sc}
\begin{tabular}{c c|c c c c c c c c}
\toprule
\multicolumn{2}{c|}{Text Datasets}  & \multicolumn{7}{c}{Method}            \\ 
$(d,[N_1, \cdots, N_K])$ & Noise Rate  & T-Rev  & CL  & HOC &  T-Vol & T-TV & Ours-$\mA$-KL & Ours-$\mA$-TV  \\ \midrule \midrule
& Low  & 10.38&	11.41&	13.32	&	9.44 & \textbf{7.08} &	{8.36}&	\textbf{8.35}  \\ 
AG's news (BERT)       & Medium  & 10.71&	10.63&	10.62	&10.53&	11.02
&	\textbf{6.44}&	\textbf{6.52 } \\ 
(768, [30k $\times$ 4]) & High  & 13.97 &	13.82&	6.80&		30.33 &	31.66
&	\textbf{4.54}&	\textbf{4.19}  \\  \midrule
               & Low  & 6.80&	5.31&	7.57&	34.23 &	34.00
&	\textbf{2.52}& \textbf{2.52}  \\ 
DBpedia (BERT)       & Medium  & 14.91&	14.40&	6.30&	26.94 &	27.62
&	\textbf{2.33}&	\textbf{2.28} \\ 
(768, [40k $\times$ 14]) & High  & 24.23&	23.28&	6.00&		30.72 &	31.35
&	\textbf{2.42}&	\textbf{2.43} \\  \midrule
               & Low  & 38.49&	38.75&	40.87&	\textbf{13.16}&	\textbf{12.46}
&	{37.37}&	{37.19 }\\ 
Yelp-5 (BERT)         & Medium  & 35.46&	36.05&	33.63&	\textbf{12.30} &	\textbf{12.48}
&	{31.79}&	{31.94}  \\ 
(768, [130k $\times$ 5]) & High  & 21.20&	20.88&	19.09&		30.09 &	33.20
&	\textbf{18.11}&	\textbf{18.06} \\  \midrule
& Low  & 20.92&	20.17&	14.25&		\textbf{3.25} &	\textbf{3.26}
&	{9.76}&	{9.97 } \\ 
Jigsaw (BERT)         & Medium  & 17.10&	16.44&	11.28&	12.23 &	12.21
&	\textbf{7.45}&	\textbf{7.66}  \\ 
(768, [144,277$,$ 15,294]) & High  & 7.19&	6.81&	4.84&		32.39 &	32.39
&	\textbf{0.78}&	\textbf{1.02} \\
\bottomrule
\end{tabular}
\end{sc}
    \end{small}
\end{center}
\vskip -0.1in %
\end{table*}

\subsection{More Experiments}\label{appendix:discuss}
\paragraph{High-noise settings may have lower errors}

Table~\ref{tab:exp2} shows the estimation error may decrease for higher noise rate settings. This observations mainly due to two reasons: 
\squishlist
\item Reason-1: The original dataset may be noisy. Notably, the original Yelp dataset contains lots of noisy reviews \cite{luca2016reviews}. Now we analyze the issues caused by an originally noise dataset with a toy example. 

\textbf{Example:} Consider a binary classification with inherent 20\% noise. Define two noise settings: \textit{1) Low noise}: Add 10\% symmetric label noise ($e_1=e_2=0.1$). \textit{2) High noise}: Add 40\% symmetric label noise ($e_1=e_2=0.4$).
For the low noise setting, the real average noise rate is:
\[
e_{\text{low, real}} = 0.1 \times 0.8 + 0.2 \times 0.9 = 0.26.
\]
For the high noise setting, the real average noise rate is:
\[
e_{\text{high, real}} = 0.4 \times 0.8 + 0.2 \times 0.6 = 0.44.
\]
Note $e_{\text{low, synthetic}} = 0.1$ and $e_{\text{high, synthetic}} = 0.4$.
Thus the gap between the real noise rate and the synthetic noise rate is
\[
e_{\text{low, real}} - e_{\text{low, synthetic}} = 0.26-0.1=0.16, \qquad 
e_{\text{high, real}} - e_{\text{high, synthetic}} = 0.44-0.4=0.04.
\]
Therefore, with inherent label noise exists, the perfectly estimated $\mT$ will have an error of $0.16$ for the low-noise setting and an error of $0.04$ for the high-noise setting, which accounts for our current observation.

\item Reason-2: The tolerance of noise rates for different settings are different. Consider a binary classification with symmetric label noise, i.e., $e_1=e_2=e$. Let the random guess be $e_1 = e_2 = 0.5$.
Define the estimation error caused by the random guess as the tolerance.
Thus the tolerances when $e=0.1$ and $e=0.4$ are $0.4$ and $0.1$, respectively.
From this aspect, an error of $0.1$ will not destroy the low-noise case since
\[
|\hat e_{1, \text{low}} - e_{1, \text{low}}| = 0.1 \Rightarrow \hat e_{1, \text{low}} =  0.2.
\]
But an error of $0.1$ may destroy the high-noise case since
\[
|\hat e_{1, \text{high}} - e_{1, \text{high}}| = 0.1 \Rightarrow \hat e_{1, \text{high}} =  0.3 \text{~~or~~} 0.5.
\]
Therefore, although the error of high-noise settings seems low, it may cause severe problems.

\squishend

\paragraph{A preliminary test on calibrating inherent errors}
We do the following experiment to help explain Reason-1 by calibrating the inherent label noise in Yelp-5.
Note the label noise accumulation follows:
\[
\mT_{\text{real}} = \mT_{\text{org}} \mT_{\text{synthetic}},
\]
where $\mT_{\text{synthetic}} = \mT$.
If we know $\mT_{\text{org}}$, we can calibrate $\mT_{\text{synthetic}}$ and evaluate the error by $\textsf{Error}(\mT_{\text{org}} \mT,\mT_{\text{org}} \hat \mT)$.
Unfortunately, we cannot find the ground-truth $\mT_{\text{org}}$ for Yelp-5. For a preliminary test, we estimate $\mT_{\text{org}}$ by applying {\sc Ours-$\mA$-KL} on the original Yelp dataset. We show the calibrated error in Table~\ref{tab:yelp_calibrate}, where we can find the high-noise settings are indeed more challenging (higher error) compared with the low-noise setting.

\begin{table*}[h]
\caption{The calibrated estimation error ($\times 100$) on Yelp-5. Average noise rate follows $e={1}/{(1+r/\sqrt{K-1})}$. {\sc Low: $r=8$. Medium: $r=4$. High: $r=1.5$.}}
\label{tab:yelp_calibrate}
\vskip 0.1in %
\begin{center}
    \begin{small}
    \begin{sc}
\scalebox{1}{
\begin{tabular}{c c|c c}
\toprule
\multicolumn{2}{c|}{Text Datasets}  & \multicolumn{2}{c}{Method}            \\ 
$(d,[N_1, \cdots, N_K])$ & Noise Rate  & Ours-$\mA$-KL & Ours-$\mA$-TV  \\ \midrule \midrule
               & Low  &	3.56&	4.01
\\ 
Yelp-5 (BERT)         & Medium  & 3.46&	2.59
  \\ 
(768, [130k $\times$ 5]) & High  & 	8.59&	8.56
 \\  
\bottomrule
\end{tabular}}
\end{sc}
    \end{small}
\end{center}
\vskip -0.1in %
\end{table*}

\end{document}